\documentclass[10pt,letterpaper]{article}

\usepackage{cvpr}           

\usepackage{url}

\usepackage{xspace}
\usepackage{dsfont}
\usepackage{afterpage}

\usepackage{enumitem}
\usepackage{booktabs}
\usepackage{arydshln}
\usepackage{caption} 
\usepackage{multirow} 
\usepackage{enumitem}
\usepackage{colortbl} 
\usepackage{bbm}
\usepackage{algorithm}
\usepackage{algpseudocode}                  
\usepackage{setspace}                       
\usepackage{lipsum} 
\usepackage{subcaption} 
\usepackage{xcolor}

\usepackage{wrapfig}
\usepackage{pifont}

\usepackage{mathtools} 
\usepackage{bm} 
\usepackage{soul} 
\usepackage{nicefrac}
\usepackage{csquotes}
\usepackage{array}
\usepackage{duckuments}

\makeatletter
\newcommand*{\myfnsymbol}[1]{%
  \ensuremath{%
    \ifcase#1\or
      \text{\ding{41}}\or
      1\or
      2\or
      3\or
      4\or
      5\or
      6\or
      7\else
      8\fi
  }%
}
\def\@fnsymbol#1{\myfnsymbol{#1}}
\def\thempfootnote{\myfnsymbol{\c@mpfootnote}}
\makeatother

\definecolor{cvprblue}{RGB}{157,49, 251}
\usepackage[pagebackref,breaklinks,colorlinks,allcolors=cvprblue]{hyperref}

\newcommand{\sname}{DMDR\xspace}

\def\paperID{-1} 
\def\confName{CVPR}
\def\confYear{2099}

\title{Distribution Matching Distillation Meets  Reinforcement Learning}

\author{
\vspace{-20pt}\\
\fontsize{11pt}{10pt}\selectfont
Dengyang Jiang$^{1,2,3}$~
Dongyang Liu$^{6,2}$ ~
Zanyi Wang$^{6}$ ~  
Qilong Wu$^{2}$  ~
Liuzhuozheng Li$^{1}$ ~\\[0.5mm]
\fontsize{11pt}{10pt}\selectfont
Hengzhuang Li$^{1}$ ~
Xin Jin$^{2}$ ~
David Liu$^{6,2}$ ~
Changsheng Lu$^{1}$ ~
Zhen Li$^{2}$ ~\\[0.5mm]
\fontsize{11pt}{10pt}\selectfont
Bo Zhang$^{4}$ ~
Mengmeng Wang$^{5}$ ~
Steven Hoi$^{2}$ ~
Peng Gao$^{2,3}$\thanks{Corresponding authors} ~
Harry Yang$^{1}$\textsuperscript{\ding{41}} \\[2mm]
\fontsize{10.2pt}{9.7pt}\selectfont
$^{1}$HKUST\hspace{1mm}
$^{2}$Alibaba Group \hspace{1mm} 
$^{3}$SIAT, CAS \hspace{1mm}
$^{4}$Shanghai AI Lab\hspace{1mm}
$^{5}$ZJUT\hspace{1mm}
$^{6}$CUHK\\[1mm]
\fontsize{10.2pt}{9.7pt}\selectfont
\textit{Code: \url{https://github.com/vvvvvjdy/dmdr}}
\vspace{-5pt} \\
}

\makeatletter
\g@addto@macro\@maketitle{
\vspace{-2em}
\begin{figure}[H]
\setlength{\linewidth}{\textwidth}
\setlength{\hsize}{\textwidth}
\centering
\includegraphics[width=\textwidth]{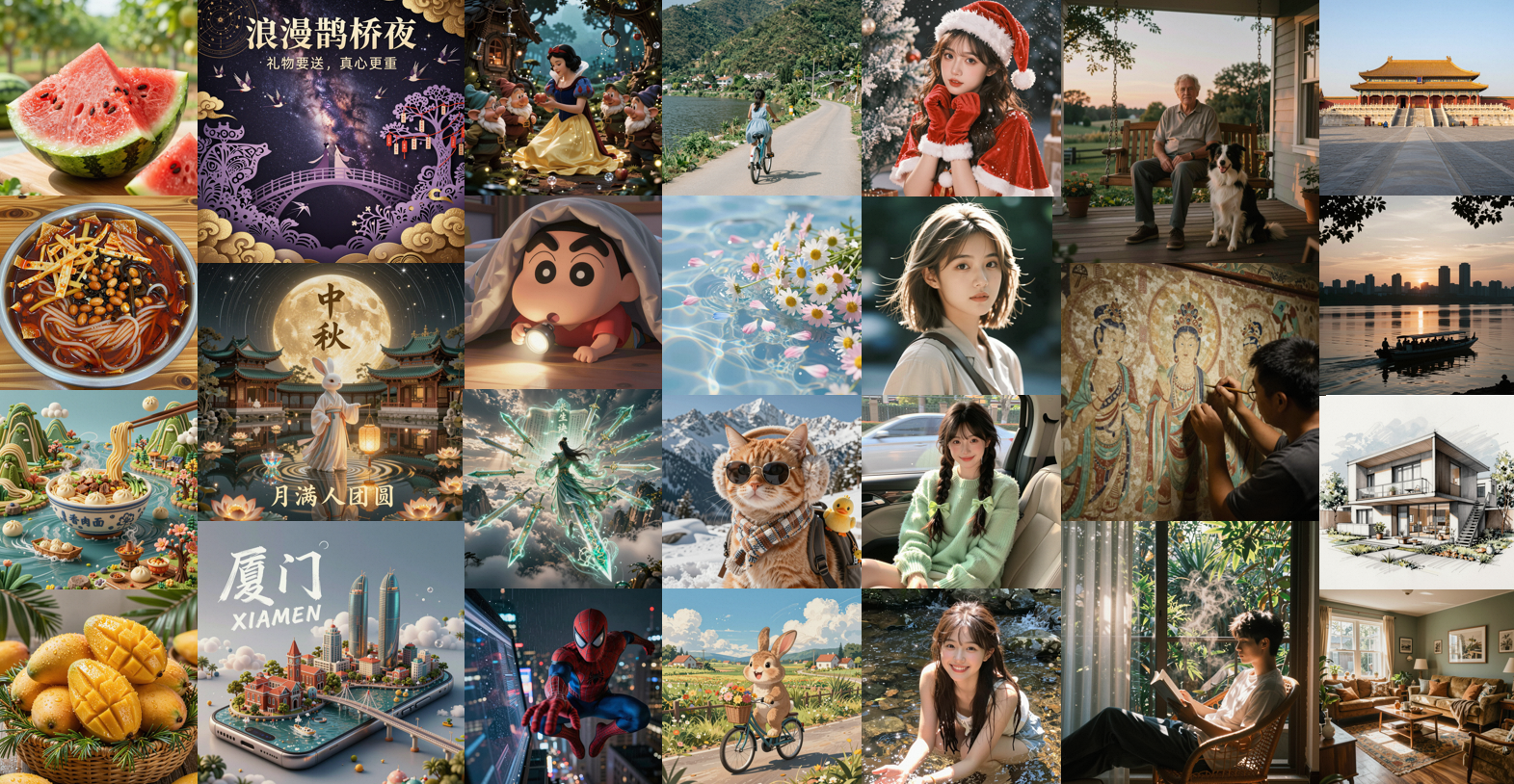}
\caption{\textbf{Images generated by Z-Image-Turbo~\cite{zimage} distilled through Decoupled-DMD~\cite{decoupled-dmd} and \sname.} Demonstrating excellent generation quality, ultra-realistic, outstanding concept understanding, and remarkable text rendering.}

\label{fig:begin_vis_z}
\end{figure}
}
\makeatother

\begin{document}
\maketitle

\begin{abstract}

Distribution Matching Distillation (DMD) facilitates efficient inference by distilling multi-step diffusion models into few-step variants. Concurrently, Reinforcement Learning (RL) has emerged as a vital tool for aligning generative models with human preferences. While both represent critical post-training stages for large-scale diffusion models, existing studies typically treat them as independent, sequential processes, leaving a systematic framework for their unification largely unexplored. In this work, we demonstrate that jointly optimizing these two objectives yields mutual benefits: RL enables more preference-aware and controllable distillation rather than uniformly compressing the full data distribution, while DMD serves as an effective regularizer to mitigate reward hacking during RL training.  Building on these insights, we propose \sname, a unified framework that incorporates Reward-Tilted Distribution Matching optimization alongside  two dynamic distillation training strategies in the initial stage, followed by the joint DMD and RL optimization in the second stage. Extensive experiments demonstrate that \sname achieves state-of-the-art visual quality and prompt adherence among few-step generation methods, even surpassing the performance of its multi-step teacher model.

\end{abstract}

\section{Introduction}
\label{sec:intro}

Diffusion models pre-trained on large-scale datasets have achieved remarkable performance in visual generation tasks~\cite{dit,sd3,flux,zimage,sdxl}. However, their sampling process typically requires numerous iterative denoising steps~\cite{ddim,ddpm,flow-matching}, each involving a full forward pass through a large neural network. This requirement makes high-resolution text-to-image synthesis computationally demanding. Furthermore, base models often fail to produce images that align closely with human preferences and aesthetic criteria~\cite{pickscore,hpsv2,hpsv3}.

To enhance practical utility, researchers typically employ two post-training stages. First, step-distillation~\cite{diffinstruct,dmd,lcm,hypersd,dmd2,flashdiff}, which compresses the original model into a generator capable of few-step sampling (e.g., 4 steps) for efficient sampling. Among such methods, Distribution Matching Distillation (DMD)~\cite{dmd} is widely recognized for its effectiveness in large-scale scenarios~\cite{dmd2,dmdx,twinflow,senseflow} and has been adopted in prominent industrial applications~\cite{seedream4,zimage,flux-2}. Second, reinforcement learning (RL)~\cite{refl,flow-dpo,flowgrpo,ddpo,dancegrpo,srpo,diffusionnft,diffusiondpo} aligns the model with human preferences to improve aesthetic quality. Despite progress in both domains, most existing studies address these objectives in isolation. Early efforts to incorporate RL into the distillation process were largely confined to a two-stage paradigm~\cite{diffinstruct++,hypersd,pso,hypernoise} (exploring how to conduct RL on an already distilled model). Consequently, there remains a critical gap in the field: \textit{a systematic framework to explore the unification and potential synergy between these two ``temporarily separated"  stages}.

To bridge this gap, we propose \sname, a unified framework that seamlessly integrates DMD and RL into a synergistic training pipeline. Our core insight is that concurrent optimization yields mutual benefits: (1) Preference-Aware Distillation: RL steers the distillation process toward high-reward regions, breaking the "performance ceiling" inherent in purely mimicking a teacher; (2) Distributional Regularization: The DMD objective serves as a robust regularizer for RL, mitigating reward hacking by maintaining proximity to the teacher's manifold.
Specifically, \sname operates in two stages. In the initial stage, we implement Reward-Tilted Distribution Matching (RT-DM), which distills toward a reward-tilted teacher distribution to prioritize human-preferred regions. Concurrently, we introduce Dynamic Distribution Guidance (DynaDG) and Dynamic Renoise Sampling (DynaRS) to stabilize the "cold-start" phase through annealed distributional overlap maximization. In the second stage, we transition to a joint optimization paradigm. In this phase,  the RL objective drives aggressive preference alignment, while the DMD mechanism serves as a robust  regularizer to prevent reward hacking.

Our experimental results show that our method leads to state-of-the-art few-step generative models. Moreover, our method is not only compatible with different models (e.g., flow-based, denoising-based) but also with various RL algorithms (e.g., ReFL, DPO, GRPO). This ensures the long-term effectiveness of our method as the multi-step model and RL algorithm evolve.

In summary, our main contributions are as follows:
\begin{itemize}[leftmargin=*,itemsep=0mm]	
		\item We propose \sname, which shows that DMD and RL can be trained simultaneously with mutual benefits. 
         \item We design a Reward-Tilted Distribution Matching and two
         dynamic distillation training strategies for better integrating RL and distillation in the initial phase.
		\item We validate our method on various models and RL algorithms, all of which achieve excellent performance.
\end{itemize}

\section{Method}

\subsection{Preliminary}
\label{sec:Pre}

\noindent\textbf{Distribution matching distillation.} DMD~\cite{dmd} compresses a multi-step diffusion model (teacher) into a few-step generator (student)~$G$ by minimizing the time-averaged approximate Kullback-Leibler (KL) divergence between the real distribution~$p_{\text{real},t}$ and the synthetic distribution~$p_{\text{fake},t}$. 
Note that $G$ is optimized via gradient descent, and this gradient admits a compact expression as the difference of two score functions:
\begin{equation}\label{eq:kl-grad}
\begin{aligned}
\nabla_{\!\theta}\,\mathcal{L}_{\text{dmd}}
&= \mathbb{E}_{t}\!\left[\nabla_{\!\theta}\operatorname{KL}\!\left(p_{\text{fake},t}\,\middle\|\,p_{\text{real},t}\right)\right] \\
&= \mathbb{E}_{t}\!\left[\nabla_{\!\theta}\mathbb{E}_{\boldsymbol{x}_0\sim p_{\text{fake},t}}
\!\left[\log\frac{p_{\text{fake},t}(\boldsymbol{x}_0)}{p_{\text{real},t}(\boldsymbol{x}_0)}\right]\right] \\
&= -\mathbb{E}_{t}\!\left[\int\!\Bigl(s_{\text{real}}\bigl(F_t\bigr)-s_{\text{fake}}\bigl(F_t\bigr)\Bigr)
   \frac{\mathrm{d}G_{\theta}(z)}{\mathrm{d}\theta}\,\mathrm{d}z\right]\!,
\end{aligned}
\end{equation}

where $z\sim\mathcal{N}(0,\mathbf{I})$, $\theta$~denotes the parameters of~$G$, and $F_t$~is the forward diffusion operator that injects noise at time~$t$ to $G_{\theta}(z)$.
The quantities $s_{\text{real}}$ and $s_{\text{fake}}$ are the score functions estimated by diffusion models $\mu_{\text{real}}$ and $\mu_{\text{fake}}$, respectively. During training, the $\mu_{\text{fake}}$ is updated via diffusion loss $\mathcal{L}_{\text{diff}}$ (for denoising-based models, the prediction target is noise,
while for flow-based ones, it is velocity) on synthetic samples produced by the few-step generator. Moreover, as introduced in DMD2~\cite{dmd2}, $G_{\theta}(z)$ is a noisy synthetic image produced by the current $G$ running several steps, which is called backward simulation. The generator $G$ then denoises these simulated images, and the
outputs are supervised with the above loss functions.

\noindent\textbf{Reinforce learning for diffusion models.}  Although there are various RL algorithms~\cite{refl, flowgrpo,diffusiondpo}, All their goal is to optimize the diffusion model to generate the samples that can maximize a score given by reward models~\cite{hpsv2,pickscore,clipscore}. This score can be expressed as aesthetic, texture, prompt following, and so on. At the same time, a regularization term is often added to balance the trade-off between reward maximization and the deviation from the original model~\cite{rlhf}. Thus, the loss can be formulated as follows:

\begin{equation}\label{eq:rl}
    \mathcal{L}_{\text{rl}} = \mathbb{E} \left[ r(\boldsymbol{x}_0) - \operatorname{KL}(p_{new}\|p_{\text{ref}}) \right],
\end{equation}
where $r$ is the reward function, $p_{new}$ and $p_{ref}$ is the distribution of the trainable model and the reference (this can be a pre-train data distribution~\cite{refl} or the original model's~\cite{flowgrpo,diffusiondpo,diffusionnft}), $\boldsymbol{x}_0$ is the clean image that is generated by the training generation model.

\subsection{Problem Formulation and Pipeline Overview.}
 It is noted that DMD effectively reduces the sampling steps of the diffusion model, while RL enables the diffusion model to generate instruction-aligned and human-preferred images. Both of these are crucial for the practical application of the diffusion model. However, at present, the majority of the work~\cite{dmdx,dmd2,magicdistillation,flowgrpo,diffusionnft,dancegrpo} only focuses on studying and improving one aspect, without exploring the synergy between the two. And only a small portion of the works~\cite{pso,flash-dmd,diffinstruct++,hypersd,hypernoise} attempt to introduce RL into the distillation process, but they still focus on how to conduct RL on a distilled model. This raises a question: \textit{Can these two be combined into one stage and benefit each other?}
 
\begin{figure*}[t]
    \centering
    \includegraphics[width=1\linewidth]{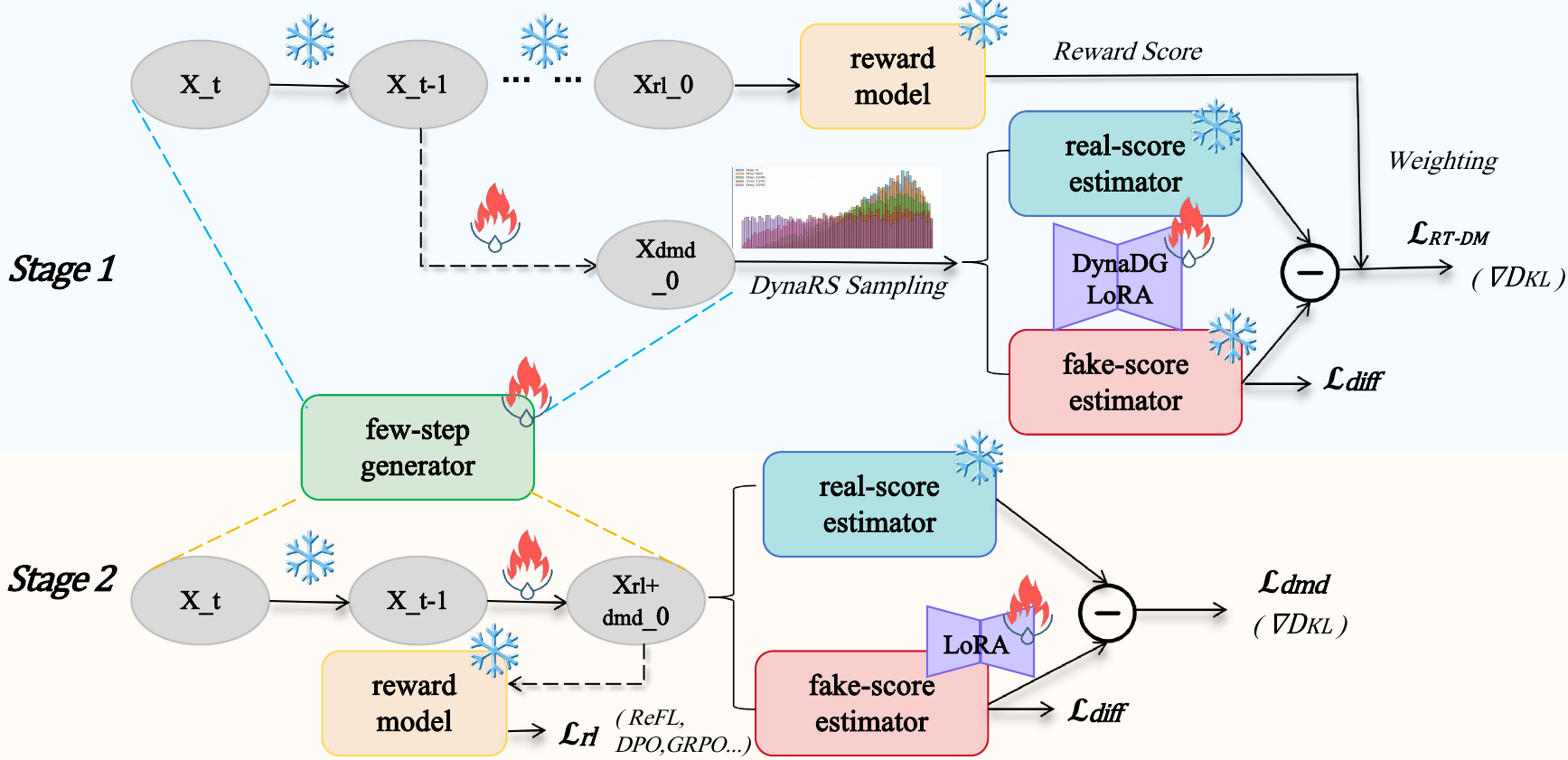}
    \caption{\textbf{Overview of \sname}. It follows a two-stage training paradigm: Stage 1 (Reward-Tilted Distribution Matching) employs reward-weighted distillation to integrate preference signals into the early distillation phase, stabilized by Dynamic Distribution Guidance (DynaDG) and Dynamic Renoise Sampling (DynaRS). Stage 2 (Joint RL + DMD) performs direct reward maximization regularized by the DMD loss.}
    \vspace{-1em}
    \label{fig:method}
\end{figure*}

 We address this challenge question by proposing \sname, as shown in Figure~\ref{fig:method}, the training process of \sname is divided into two operational stages: Stage 1: RT-DM with Dynamic Distillation. (Section~\ref{sec:soft-rl} and Section~\ref{sec:dyna}) Stage 2: Joint RL + DMD Optimization(Section~\ref{sec:s2dmdr}). The following sections will provide the detailed analysis and illustration.  

\subsection{Bring DMD and RL Together with Benefit}
\label{sec:s2dmdr}

We now show formally and empirically that DMD operates as a superior regularization mechanism that subsumes the traditional KL penalty in RL for few-step model, yielding robust joint optimization.

\begin{wrapfigure}{r}{0.4\textwidth}
    \centering
    \vspace{-1em}
    \includegraphics[width=1\linewidth]{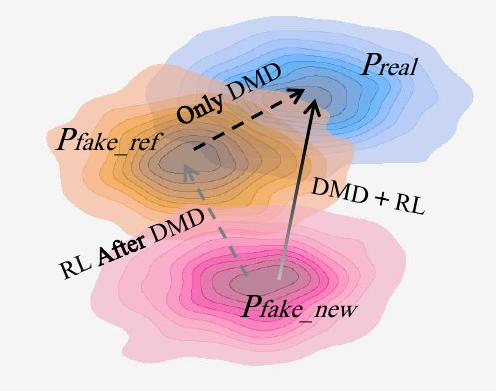}
    \caption{Visualization of the difference of optimization directions.}
    \label{fig:bene_to_rl}
    \vspace{-2em}
\end{wrapfigure}

From the perspective of RL, standard preference optimization requires a penalty (e.g., $\operatorname{KL}(p_{new}\|p_{\text{ref}})$) to prevent reward hacking~\cite{refl, flowgrpo}. In few-step models, sequentially applying RL after distillation uses the already-distilled distribution $P_{\text{fake-ref}}$ as the regularizer (Fig.~\ref{fig:bene_to_rl}). Because distillation inherently loses some modes, $P_{\text{fake-ref}}$ acts as an impoverished anchor, leading to rapid overfitting and mode collapse~\cite{hypersd}. 
More fundamentally, instead of viewing DMD and RL as two disparate losses added heuristically, we reframe the joint objective as maximizing reward under a distribution-level structural constraint.  As the joint gradient can be decomposed into two orthogonal objectives~\cite{rlhf,score-base-diff,ppo}: a "reward ascent" direction that increases human preference, and a "manifold projection" term that corrects off-manifold displacement by pulling the student back toward the teacher’s  regions.The DMD gradient $\nabla_{\!\theta}\mathcal{L}_{\text{dmd}}$ corresponds exactly to the maximum likelihood update projecting the student output distribution toward the learned data distribution of the multi-step teacher $P_{\text{real}}$. When combined linearly with the reward gradient $\nabla_{\!\theta}\mathcal{L}_{\text{rl}}$, any displacement induced by $r$ that exits the teacher's high-density manifold generates a large discrepancy $s_{\text{real}} - s_{\text{fake}}$. The $\mathcal{L}_{\text{dmd}}$ gradient thus projects the reward ascent strictly along the teacher’s support.

Simultaneously, from the perspective of distillation, pure DMD forces the student to uniformly mimic the teacher, implicitly establishing a "performance ceiling" capped by distilling $P_{\text{real}}$ equally. Incorporating the RL objective addresses this fundamental limitation: RL steers the distillation  toward high-reward regions, allowing the student to surpass the teacher while the DMD regularizer ensures semantic integrity is preserved.

\begin{figure*}[t]
    \centering
    \includegraphics[width=1\linewidth]{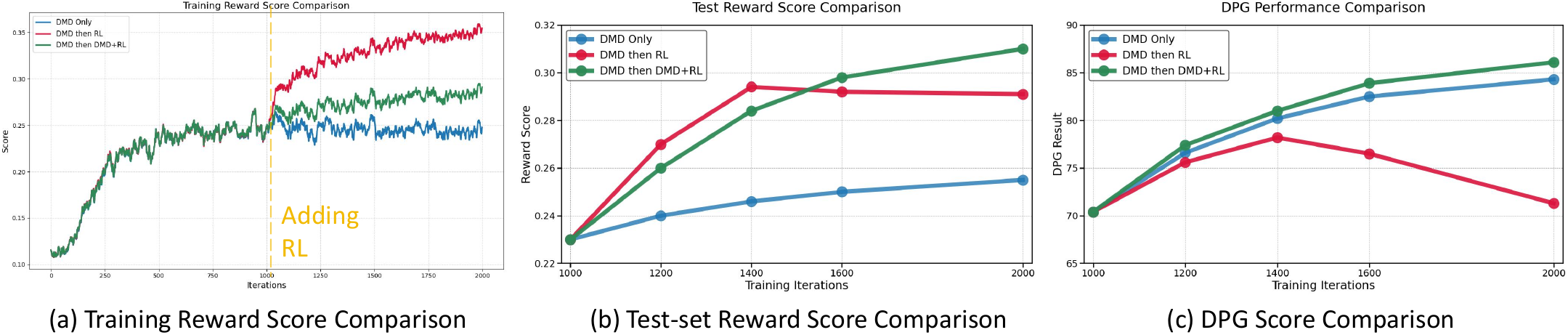}
    \caption{\textbf{Verification of the joint optimization synergy.} Directly conduct RL (red) on the distilled model exhibits high training rewards but suffers from reward hacking and semantic collapse (low DPG score). In contrast, the joint DMD+RL paradigm (green) effectively boosts the performance (better than the DMD baseline (blue) in all metric) while preventing reward hacking. Better to
zoom in to check the effect.}
    \vspace{-1em}
    \label{fig:verification1}
\end{figure*}
  \begin{wrapfigure}[14]{r}{0.66\textwidth}
\centering\small
\captionof{table}{ 
\textbf{Comparison of different training strategies.} The joint approach avoids the "distillation gap" of "RL then distillation" pipeline, achieving the highest preference score with $\sim 10 \times$ rollout speedup.
}
\resizebox{0.64\textwidth}{!}{
\begin{tabular}{lccc}
     \toprule 
     Method & HPS &  DPG & Time\\ 
     \midrule
  \multicolumn{4}{l}{\emph{SD3-M, resolution 1024$\times$1024, local batchsize 16 on H100}} \\
  \midrule
     Multi-Step RL& 31.06 & 86.43 & 9.24s $\sim$ 10.57s \\
     Multi-Step RL then Distillation& 30.42 & 85.92 & - \\
     Few-Step Joint RL and DMD & 31.42 & 86.08 & 0.43s $\sim$ 1.21s \\
     \bottomrule 
\end{tabular}
}
\label{tab:verification2}
\end{wrapfigure}

 To validate the analysis, we conduct a comparative study using SD3-M~\cite{sd3} as the base model, with rewards provided by HPS v2.1~\cite{hpsv2} and ReFL-based RL~\cite{refl} for training. We compare three paradigms: (i) DMD Only (Blue line); (ii) DMD then RL (Sequential: distillation followed by RL in red line); and (iii) DMD then DMD+RL (Joint approach: distillation followed by joint optimization in green line). We evaluate these models across training/test reward curves and the DPG\_Bench  score~\cite{dpg}.
 As shown in Fig.~\ref{fig:verification1}(a), when RL is introduced (after 1000 iterations), the sequential approach achieves the highest training reward. However, this gain is obtained through hacking score. Fig.~\ref{fig:verification1}(b) and (c) reveal that this it suffers from severe reward hacking and catastrophic forgetting. While its training reward climbs, its test reward plateaus, and its DPG score collapses. This confirms that without the strong distributional anchor of the teacher, the few-step model quickly drifts into narrow, high-reward modes that sacrifice semantic integrity~\cite{hypersd}.
In contrast, the joint optimization demonstrates a clear synergistic effect. First, it effectively breaks the "performance ceiling" of the teacher; the test reward significantly surpasses the DMD-only baseline. Second, by treating the DMD loss as a persistent structural regularizer, the model keeps improving DPG score while increasing aesthetic quality. This indicates that the DMD objective prevents the RL process from hacking, while the RL objective guides the distillation toward more human-preferred regions.

Based on the empirical synergy observed above, we formally define the optimization objective. The total loss is formulated as:
\begin{equation}\label{eq:stage2-loss}
\mathcal{L}_{\text{total}} = \mathcal{L}_{\text{dmd}} + \lambda_{\text{rl}} \mathcal{L}_{\text{rl}},
\end{equation}
where $\mathcal{L}_{\text{dmd}}$  is the original distribution matching loss with the gradient formulated in Eq.~\ref{eq:kl-grad}, and $\mathcal{L}_{\text{rl}}$ is the plug-and-play reward maximization loss from the RL branch (Eq.~\ref{eq:kl-grad}), the
gradient can be differentiable type like ReFL, or policy type
like GRPO, etc. The hyperparameter $\lambda_{\text{rl}}$ is a balancing coefficient used to trade off the two losses.

\subsection{Reward-Tilted Distribution Matching for Integrating RL into the Early Phase of DMD}
\label{sec:soft-rl}
While we have demonstrated that DMD and RL can be trained jointly with mutual benefit, we still rely on a two-stage paradigm that incorporates RL only after an initial "cold start" distillation phase~\cite{hypersd,diffinstruct++}. This requirement stems from the observation that reward models typically necessitate a baseline generation capability to produce reliable feedback \cite{refl, srpo}. Specifically, in DMD frameworks with flow matching model, the image  used for gradient computation is defined as 
$\boldsymbol{x}_t - (0 - t) G_{\theta}(\boldsymbol{x}_t)$, where $\boldsymbol{x}_t$ is iteratively sampled.  Notably, t can be any training time-step; thus, when training a 4-step model, the clean image can be generated in only one to four steps. In the early training phase, however, such few-step sampling often fails to produce coherent images, rendering reward signals noisy and unreliable for direct optimization.

\begin{figure*}[t]
    \centering
    \includegraphics[width=1\linewidth]{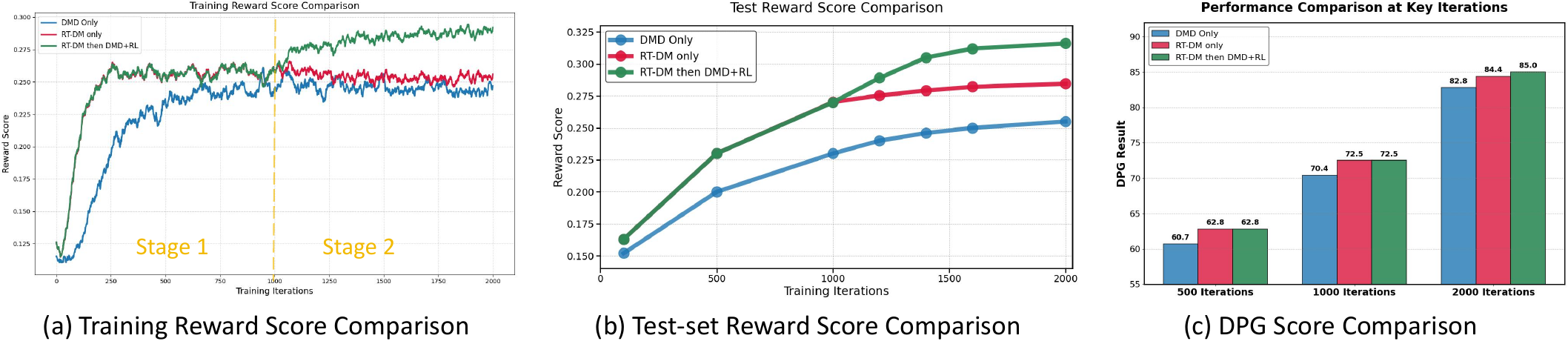}
    \caption{\textbf{Effectiveness of RT-DM and two-stage training strategy.} In Stage 1, the RT-DM paradigm (red/green) significantly accelerates preference alignment compared to the DMD baseline (blue). In Stage 2, while "RT-DM only" (red) eventually saturates due to the constraints of the teacher's manifold, the transition to joint DMD+RL (green) successfully breaks this performance plateau by directly optimizing the reward score. This complete two-stage approach achieves the highest test rewards and DPG scores. Better to zoom in to check the effect.}
    \vspace{-1em}
    \label{fig:verification3}
\end{figure*}

To address this, we propose a Reward-Tilted Distribution Matching (RT-DM) paradigm to "softly" integrate reward signals into the distillation process. Specifically, we utilize the reward score as a weighting factor for the DMD loss. This formulation enables a more flexible optimization objective compared to direct reward maximization. This is because the reward score serves as a static weighting factor, we avoid the necessity of backpropagating through the multi-step denoising process, which allows us to evaluate the reward on the complete trajectory of the few-step model (e.g., 4 steps) without the computational overhead of a large gradient graph, bypassing the need for single-step $\boldsymbol{x}_0$ approximations to   make the reward score more precise and highly valuable as a reference for optimization efforts. More profoundly, we are changing \textit{which} distribution the student learns by assigning larger gradients to regions with higher reward scores, so that the model is not merely encouraged to match the real distribution defined by the teacher, but is instead biased toward better aligning with regions of the teacher distribution that yield higher reward scores.
The gradient is formulated as:

\begin{equation}\label{eq:kl-grad-rl}
\begin{aligned}
\nabla_{\!\theta}\,\mathcal{L}_{\text{RT-DM}}
&= \mathbb{E}_{t}\!\left[\nabla_{\!\theta}\Bigl(\operatorname{KL}\!\left(p_{\text{fake},t}\,\middle\|\,p_{\text{real},t}\right)\,R(\boldsymbol{x}_0)\Bigr)\right] \\
&= \mathbb{E}_{t}\!\left[\nabla_{\!\theta}\mathbb{E}_{\boldsymbol{x}_0\sim p_{\text{fake},t}}
\!\left[R(\boldsymbol{x}_0)\log\frac{p_{\text{fake},t}(\boldsymbol{x}_0)}{p_{\text{real},t}(\boldsymbol{x}_0)}\right]\right] \\
&= -\mathbb{E}_{t}\!\left[\int\! R(\boldsymbol{x}_0)\Bigl(s_{\text{real}}\bigl(F_t\bigr)
   - s_{\text{fake}}\bigl(F_t\bigr)\Bigr)
   \frac{\mathrm{d}G_{\theta}(z)}{\mathrm{d}\theta}\,\mathrm{d}z\right]\!,
\end{aligned}
\end{equation}
where $R(\boldsymbol{x}_0) = e^{\frac{r(\boldsymbol{x}_0)}{\beta}}$, In our experiments, we set
$\beta = \max_{\boldsymbol{x}_0' \in \mathcal{B}} r(\boldsymbol{x}_0')$,
the maximum reward within the current mini-batch, to control the scale of the reward-induced weighting and prevent excessively large gradients.

To evaluate the effectiveness of the proposed soft integration, we compare three configurations in Fig.~\ref{fig:verification3}: (i) DMD Only (Blue); (ii) RT-DM for the entire duration (Red); and (iii) RT-DM followed by joint DMD+RL (Green). As shown in Fig.~\ref{fig:verification3}(a) and (b), the RT-DM paradigm (Red/Green) provides a significant "warm start" effect, accelerating preference alignment in the early phase (Stage 1) compared to the DMD-only baseline. This confirms that weighting the DMD loss with reward scores successfully incorporates preference signals even when few-step generations are still maturing, avoiding the noise instability of direct RL.
However, the performance of the "RT-DM only" approach (Red) eventually saturates. We hypothesize that since RT-DM primarily acts by re-weighting the teacher’s distribution gradients, its ability to drive the model toward high-reward regions is inherently constrained by the teacher's manifold. To further unlock the model's potential, we introduce a more direct reward maximization phase (Green) in Stage 2. By transitioning to joint DMD+RL optimization once the model has reached a stable baseline, we can leverage direct RL gradients to aggressively push the performance beyond the initial plateau. As a result, the joint approach achieves the highest test rewards and DPG scores, demonstrating that RT-DM serves as an ideal bridge to transition from pure distillation to direct preference alignment.

\subsection{Annealed Distributional Overlap Maximization for Better Initial Distillation}
\label{sec:dyna}

During the initial phase of distillation, DMD encounters a classic "cold start" challenge: the synthetic samples generated by the few-step student exhibit substantial divergence from the real-world pre-training samples. This profound disjointness between $P_{\text{fake}}$ and $P_{\text{real}}$ impedes the teacher model’s ability to estimate reliable scores, rendering the gradient $\nabla \log P_{\text{real}} - \nabla \log P_{\text{fake}}$ unreliable for optimization. To mitigate this, we introduce a unified principle of \textit{annealed distributional overlap maximization}. We propose two complementary techniques to artificially increase the overlap between $P_{\text{fake}}$ and $P_{\text{real}}$ early in training, progressively relaxing them as the model converges.

\begin{wrapfigure}{r}{0.64\textwidth}
    \centering
    \vspace{-1em}
    \includegraphics[width=1\linewidth]{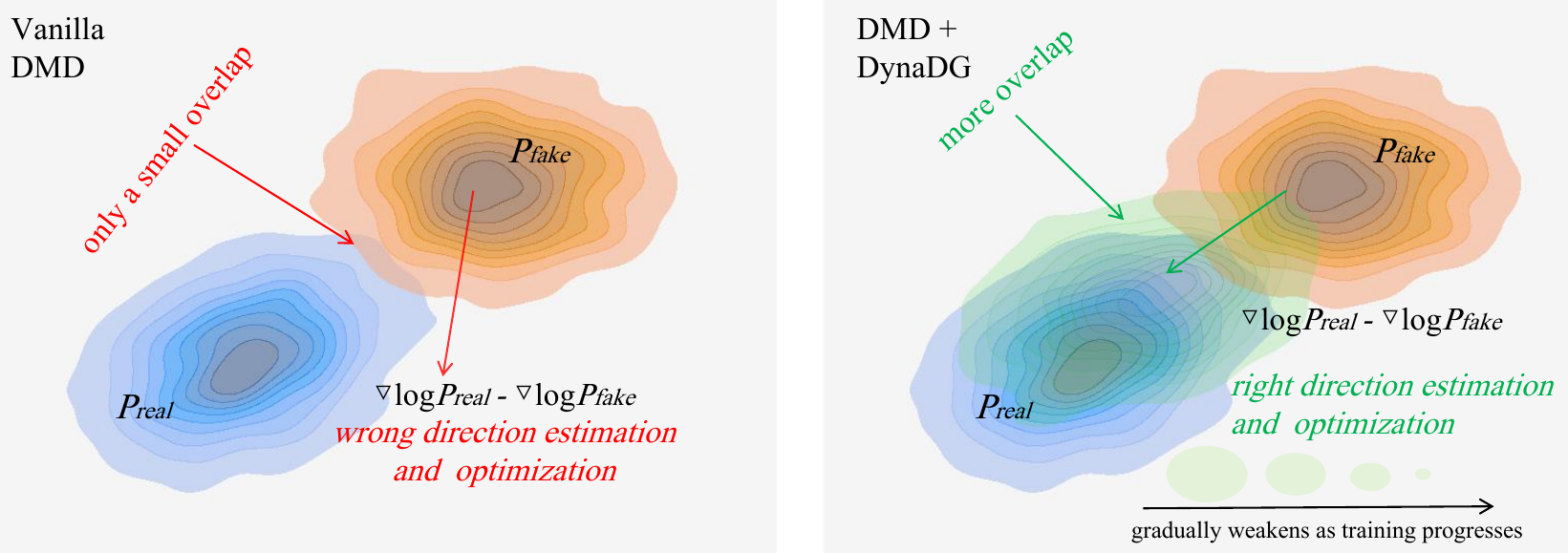}
    \caption{Illustration for Dynamic Distribution Guidance.}
    \label{fig:dynadg}
    \vspace{-1em}
\end{wrapfigure}

\noindent\textbf{Dynamic Distribution Guidance (DynaDG).} DynaDG maximizes overlap primarily by actively shifting the target distribution. The poor initial Image quality causes $P_{\text{fake}}$ to be disjoint from $P_{\text{real}}$ (Figure~\ref{fig:dynadg}, left). To solve this, we inject trainable LoRA modules~\cite{lora} into the real score estimator to "pull" the perceived $P_{\text{real}}$ toward the nascent $P_{\text{fake}}$ (Figure~\ref{fig:dynadg}, right). This ensures a reliable optimization signal by bridging the manifold gap. As training progresses, we anneal the LoRA scale in the teacher model toward zero, allowing the real score estimator to seamlessly revert to the true $P_{\text{real}}$ once $P_{\text{fake}}$ is safely anchored.

\begin{figure*}[t]
    \centering
    \includegraphics[width=1\linewidth]{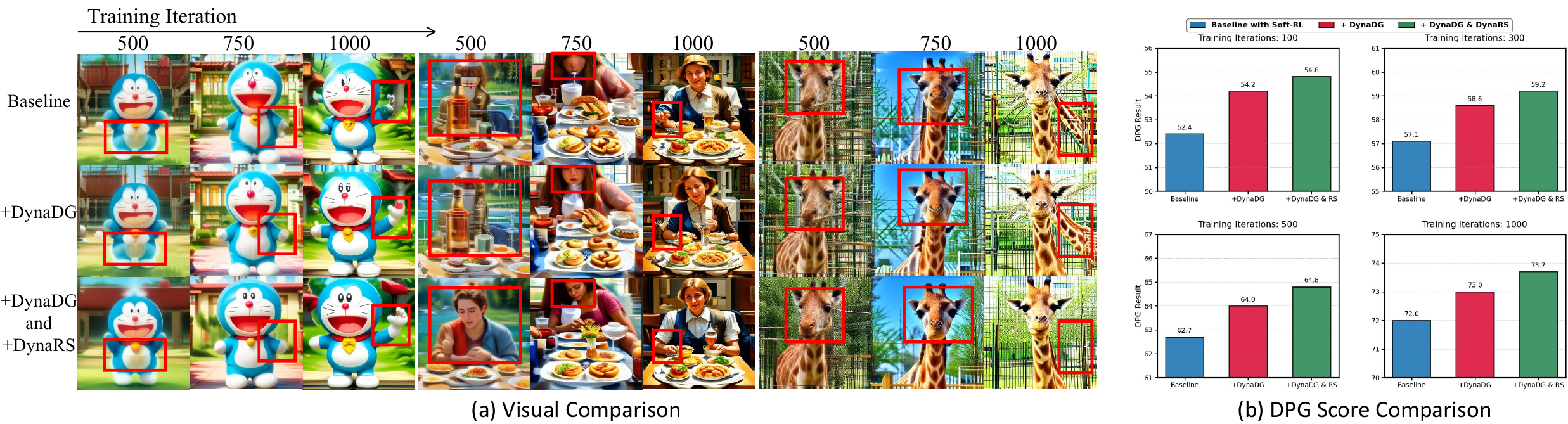}
    \caption{\textbf{Impact of dynamic training strategies}. (a) Qualitative comparisons show that our dynamic strategies facilitate faster global structure building (highlighted in red boxes). (b) Quantitative DPG scores further confirm that DynaDG and DynaRS provide more reliable optimization signals, consistently outperforming the baseline throughout Stage 1. Better to zoom in to check the effect.}
    \vspace{-1em}
    \label{fig:verification4}
\end{figure*}

\noindent\textbf{Dynamic Renoise Sampling (DynaRS).} While DynaDG moves the distributions, DynaRS maximizes overlap by exploiting regions where distributions naturally intersect. By heavily biasing the sampling of renoise levels $t$ toward higher noise values at the start of training, we ensure that both initial samples and reference samples are dominated by Gaussian noise, placing them in a similar regime. At these high noise levels, the score estimator provides highly reliable gradients that emphasize global structure~\cite{lsast,prospect}. As the student generator improves, the renoise bias is progressively annealed to uniform sampling, allowing the model to capture fine-grained textural details from a wider range of noise levels.

As shown in Figure~\ref{fig:verification4} (b),  our dynamic training strategies significantly enhance the initial training phase.  Qualitative results presented in Figure~\ref{fig:verification4} left further corroborate these findings, showing a faster global structure building when DynaDG and DynaRS are employed.

\section{Experiments}

\subsection{Experimental Setup}

For training, we adopt denoising-based model (SDXL-Base~\cite{sdxl}) and flow-based models (SD3-Medium~\cite{sd3}, SD3.5-Large~\cite{sd35}) for distillation, using prompts from t2i-2M. Additionally, we use ReFL~\cite{refl} with DFN-CLIP~\cite{dfnclip} and HPSv2.1~\cite{hpsv2} as reward models by default. As for evaluation, we validate the effect of \sname through extensive experiments. First, we use prompts sampled from a recent public dataset ShareGPT-4o-Image~\cite{sharegpt4o-image} to generate images and follow previous work~\cite{dmdx,tdm} that report CLIP Score~\cite{clipscore}, Aesthetic Score~\cite{laion5b}, Pick Score~\cite{pickscore}, and Human Preference (HP) Score~\cite{hpsv2}. Next, in order to obtain a more comprehensive evaluation, we also compare our distilled models with their teachers on two popular benchmarks, DPG\_Bench~\cite{dpg} and GenEval~\cite{geneval}. Noted that there are some differences from the settings we made in the Method section. We train 1.5K steps in stage1 and stage 2, respectively, to achieve better performance.

\begin{table*}[t]
\centering
\captionof{table}{\textbf{System-level comparison} against state-of-the-art methods. $^{\star}$ denotes our reproduced results based on the official code. Best performance is marked in \textbf{Bold}.}
\vspace{-0.5em}
\resizebox{1\linewidth}{!}{
\begin{tabular}{l c c c c c c}
\toprule
{Method} & NFE & Post-Train & CLIP Score$\uparrow$ & Aesthetic Score$\uparrow$ &Pick Score$\uparrow$ & HP Score$\uparrow$\\
\arrayrulecolor{black}\midrule

\multicolumn{6}{l}{\textcolor{gray}{\emph{SDXL-Base}}} \\
Base-Model  & 50 & - & 34.7588 & 5.6480 & 22.1085 & 27.1477 \\
ReFL~\cite{refl}$^{\star}$ & 50 & RL & 35.2107 & 5.9045 & 22.8076 & \textbf{33.1874} \\
LCM~\cite{lcm}  & 2 & Distill & 28.4664 & 5.1026 & 20.0603 & 17.6837 \\
Lightning~\cite{sdxl-lightning} & 1 & Distill & 32.0283 & 5.6761 & 21.4868 & 26.3615 \\
DMD2~\cite{dmd2} & 1 & Distill & 34.3046 & 5.6238 & 21.7293 & 26.5986 \\
\rowcolor[RGB]{240,230,245}
\sname (ours) & 1  & Distill \& RL & \textbf{35.6241} & \textbf{6.1184} & \textbf{22.8498} & 32.0065 \\

\arrayrulecolor{black!20}\midrule
DMD2~\cite{dmd2} & 4 & Distill & 34.5169 & 5.7043 & 22.1546 & 28.5655 \\
DMD2-PSO~\cite{pso}$^{\star}$ & 4 & Distill then RL & 34.0128 & 5.8032 & 22.2644 & 29.8732 \\
\rowcolor[RGB]{240,230,245}
\sname (ours) & 4 & Distill \& RL & \textbf{35.4940} & \textbf{6.0324} & \textbf{22.7122} & 32.9832 \\

\arrayrulecolor{black!40}\midrule

\multicolumn{6}{l}{\textcolor{gray}{\emph{SD3-Medium}}} \\
Base-Model  & 50 & - & 34.9025 & 5.5942 & 22.1801 & 28.4021 \\
ReFL~\cite{refl}$^{\star}$ & 50 & RL &  \textbf{35.1851} & 5.7821 & 22.0064 & 32.0745 \\
DMD2~\cite{dmd2}$^{\star}$ & 4 & Distill & 33.9421 & 5.6137 & 21.7688 & 27.3675 \\
Flash~\cite{flashdiff} & 4  & Distill & 34.2634 & 5.6702 & 21.5921 & 26.6542 \\
TDM~\cite{tdm} & 4 & Distill & 34.0301& 5.6250 & 22.0010 & 27.7522 \\
Hyper-SD~\cite{hypersd}  & 8 & Distill then RL & 32.0234 & 5.2489 & 20.2831 & 22.4544 \\
\rowcolor[RGB]{240,230,245}
\sname (ours) & 4 & Distill \& RL &35.0142 & \textbf{5.8876} &  \textbf{22.4892} & \textbf{32.1145} \\

\arrayrulecolor{black!40}\midrule
\multicolumn{6}{l}{\textcolor{gray}{\emph{SD3.5-Large}}} \\
Base-Model & 50 & - & 35.5509 & 5.7014& 22.4856 & 28.8135 \\
LADD~\cite{ladd}  & 4 & Distill then RL & 35.0480 & 5.4514 & 22.2451 & 27.8470 \\
\rowcolor[RGB]{240,230,245}
\sname (ours) & 4 & Distill \& RL & \textbf{35.9757} & \textbf{6.1541} &  \textbf{22.9072} & \textbf{32.7368} \\

\arrayrulecolor{black}\bottomrule
\end{tabular}}
\vspace{-1em}
\label{tab:system_compare}
\end{table*}

\noindent\textbf{Method for comparison.} We compare our  method against several categories of existing work, including foundational multi-step base models~\cite{sdxl,sd3,sd35}, RL only approaches~\cite{refl}, distillation only methods~\cite{lcm,sdxl-lightning,dmd2,flashdiff,tdm}, distillation then RL frameworks~\cite{hypersd,pso}. We do not compare with methods like Flash-DMD~\cite{flash-dmd} and Diff-Instruct++~\cite{diffinstruct++} that do not provide model weights or open source training code because we cannot guarantee a fair and reproducible evaluation.

\subsection{Main Results}
\begin{figure*}[t]
    \centering
    \includegraphics[width=1\linewidth]{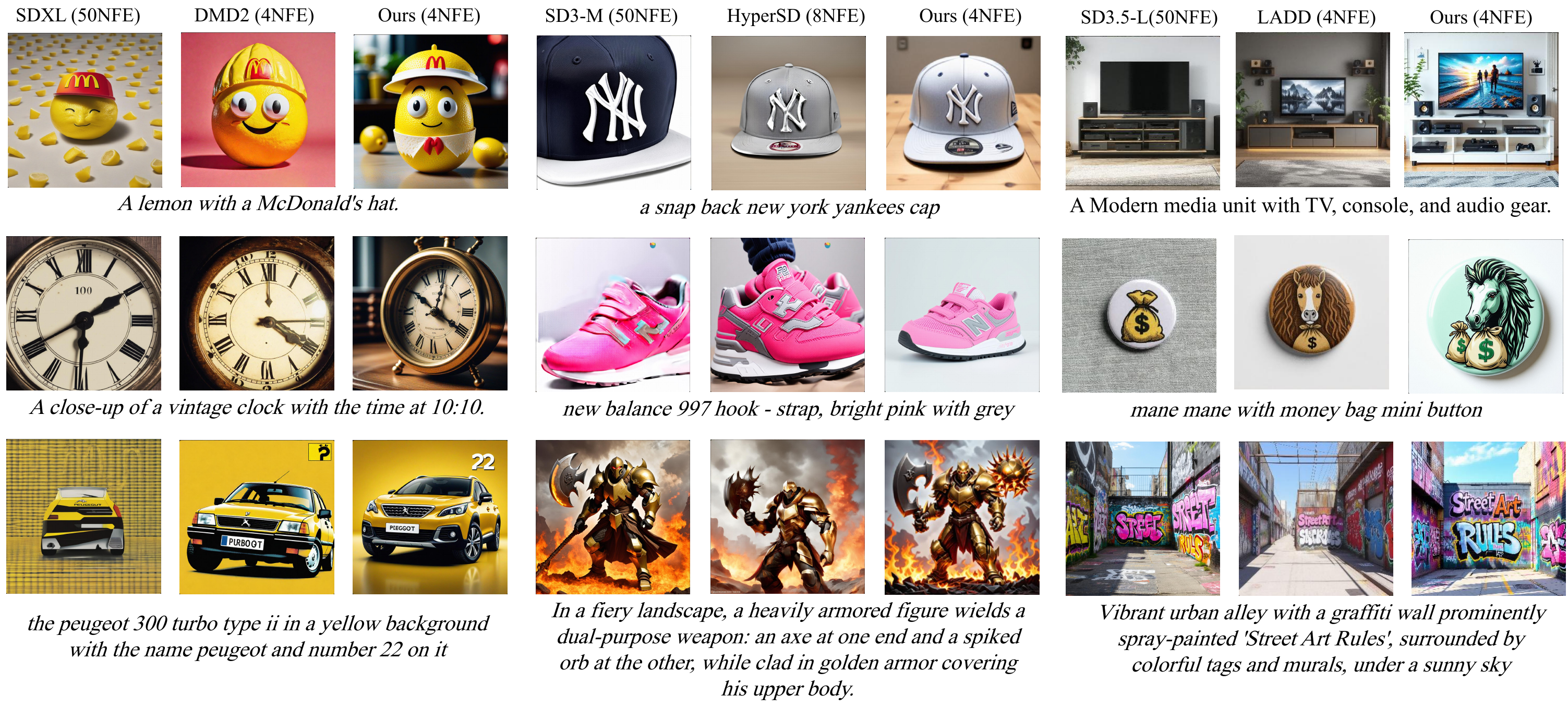}
    \caption{Visual comparison between the teachers, selected competing
methods~\cite{dmd2,hypersd,ladd}, and ours. All images are generated using identical noise. Our model produces images with superior quality and prompt coherence. }
\vspace{-1em}
    \label{fig:vis_comp_all}
\end{figure*}

We begin by assessing \sname's text-to-image generation performance using prompts sampled from ShareGPT-4o-Image~\cite{sharegpt4o-image}, ensuring that there is no overlap with the prompts utilized during training. Table~\ref{tab:system_compare} presents a summary of the distillation results in comparison to other methods. Models distilled through our approach exhibit state-of-the-art (SOTA) capabilities in terms of prompt coherence and the generation of high-quality, aesthetically pleasing images. Notably, our method demonstrates effectiveness across various architectures (e.g., UNet, Transformer), model sizes (e.g., 2B, 8B), and paradigms (e.g., flow-based, denoising-based), highlighting the broad applicability and universality of the proposed approach. Additionally, we provide qualitative comparisons in Figure~\ref{fig:vis_comp_all}, where our method produces images characterized by superior quality and enhanced prompt coherence.

\begin{table}[t]
\centering
\begin{minipage}[t]{0.49\linewidth}
\centering
\caption{\textbf{GenEval Comparison:} 4-step (ours) $vs.$ its multi-step teacher.}
\vspace{-0.5em}
\label{tab:GenEval_Bench}
\resizebox{\linewidth}{!}{
\begin{tabular}{l|c|cccccc}
\toprule
Model & Overall & Single & Two & Count. & Colors & Pos. & Attr. \\
\midrule
\multicolumn{8}{l}{\textcolor{gray}{\emph{SDXL-Base}}} \\
Teacher & 0.55 & 0.98 & 0.74 & 0.39 & 0.85 & 0.15 & 0.23 \\
Ours    & \textbf{0.57} & \textbf{0.99} & \textbf{0.76} & \textbf{0.44} & \textbf{0.85} & 0.12 & \textbf{0.25} \\
\arrayrulecolor{black!40}\midrule
\multicolumn{8}{l}{\textcolor{gray}{\emph{SD3-Medium}}} \\
Teacher & 0.62 & 0.98 & 0.74 & 0.63 & 0.67 & 0.34 & 0.36 \\
Ours    & \textbf{0.65} & \textbf{0.99} & \textbf{0.84} & 0.57 & \textbf{0.81} & 0.27 & \textbf{0.45} \\
\arrayrulecolor{black!40}\midrule
\multicolumn{8}{l}{\textcolor{gray}{\emph{SD3.5-Large}}} \\
Teacher & 0.71 & 0.98 & 0.89 & 0.73 & 0.83 & 0.34 & 0.47 \\
Ours    & \textbf{0.72} & \textbf{0.99} & \textbf{0.93} & 0.69 & 0.80 & 0.31& \textbf{0.59} \\
\arrayrulecolor{black}\bottomrule
\end{tabular}}
\end{minipage}
\hfill
\begin{minipage}[t]{0.49\linewidth}
\centering
\caption{\textbf{DPG\_Bench Comparison:} 4-step (ours) $vs.$ its multi-step teacher.}
\vspace{-0.5em}
\label{tab:DPG_Bench}
\resizebox{\linewidth}{!}{
\begin{tabular}{l|c|ccccc}
\toprule
Model & Overall & Global & Entity & Attribute & Relation & Other \\
\midrule
\multicolumn{7}{l}{\textcolor{gray}{\emph{SDXL-Base}}} \\
Teacher & 74.65 & 83.27 & 82.43 & 80.91 & 86.76 & 80.41 \\
Ours & \textbf{76.48} & \textbf{83.72} & \textbf{82.58} & \textbf{83.69} & 84.76 & \textbf{83.54} \\
\arrayrulecolor{black!40}\midrule
\multicolumn{7}{l}{\textcolor{gray}{\emph{SD3-Medium}}} \\
Teacher & 84.08 & 87.90 & 91.01 & 88.83 & 80.70 & 88.68 \\
Ours & \textbf{85.30} & \textbf{90.49} & 90.07 & \textbf{90.56} & \textbf{87.17} & \textbf{91.23} \\
\arrayrulecolor{black!40}\midrule
\multicolumn{7}{l}{\textcolor{gray}{\emph{SD3.5-Large}}} \\
Teacher & 84.12 & 91.48 & 90.22 & 87.81 & 91.20 & 89.49 \\
Ours & \textbf{85.33} & 90.47 & \textbf{90.53} & \textbf{90.68} & 87.44 & \textbf{90.20} \\
\arrayrulecolor{black}\bottomrule
\end{tabular}}
\end{minipage}
\end{table}

To further verify our method’s effectiveness and provide a more comprehensive comparison between the few-step distillation model and the multi-step teacher model, we conduct the comparison on two commonly used benchmarks~\cite{dpg,geneval}, 
Although we do not  perform RL optimization on the metric of these two benchmarks,
table~\ref{tab:DPG_Bench} and ~\ref{tab:GenEval_Bench} show that the overall score of our four-step model consistently outperforms its multi-step teacher on DPG\_Bench and GenEvalacross all base models.
 These results indicate that \sname has successfully freed the student model from the constraints of the multi-step teacher model and stimulated its capabilities during the distillation process.

\begin{table}[t]
\centering
\caption{Ablation  on different components. Default settings are marked in purple.}
\label{tab:combined_ablations}
\vspace{-1em} 
\begin{subtable}{\linewidth}
    \centering
    \caption{Ablation on different reward models.}
    \vspace{-0.5em}
    \resizebox{0.6\linewidth}{!}{ 
    \begin{tabular}{l c c c c c }
    \toprule
    \multirow{2}{*}{Method} & CLIP & Aesthetic & Pick & HP & DPG\\
      &Score & Score & Score & Score & overall\\
    \arrayrulecolor{black!40}\midrule
    Pick~\cite{pickscore} & 34.3403 & 5.9021 & 22.9674 & 31.7665 & 84.86 \\
    AE~\cite{laion5b} & 34.1748 & 6.1024 & 22.3858 & 31.6620 & 84.61 \\
    CLIP~\cite{clipscore} & 35.2087 & 5.5832 & 22.0644 & 28.0507 & 85.10 \\
    HPS~\cite{hpsv2} & 34.4088 & 5.9021 & 22.5108 & 32.9065 & 85.04 \\
    \rowcolor[RGB]{248, 242, 253} 
    CLIP + HPS & 35.0142  & 5.8876  &  22.4892 & 32.1145 & 85.30\\
    \arrayrulecolor{black}\bottomrule
    \end{tabular}}
    \label{tab:ab_reward_model}
\end{subtable}

\vspace{1em} 

\begin{subtable}{0.52\linewidth}
    \centering
    \caption{Ablation on RL algorithm in stage 2.}
    \vspace{-0.5em}
    \resizebox{\linewidth}{!}{
    \begin{tabular}{l c c  c c c }
    \toprule
    \multirow{2}{*}{Method} & CLIP & Aesthetic & Pick & HP & DPG\\
      &Score & Score & Score & Score & overall\\
    \midrule
    {\textcolor{gray}{\emph{stage 1 init}}} & {\textcolor{gray}{\emph{33.3648}}} & {\textcolor{gray}{\emph{5.5744}}}
    & {\textcolor{gray}{\emph{21.0070}}} & {\textcolor{gray}{\emph{28.8371}}} & {\textcolor{gray}{\emph{82.4371}}}\\
    \arrayrulecolor{black!40}\midrule
    \rowcolor[RGB]{248, 242, 253} 
    w/  RL (ReFL) & 35.0142  & 5.8876  &  22.4892 & 32.1145 & 85.30\\
    w/  RL (DPO) & 34.0234 & 5.8340 & 21.9844 & 30.8352 & 85.00 \\
    w/  RL (GRPO) & 34.6675 & 5.9324 & 22.3248 & 31.4474 & 85.08\\
    \arrayrulecolor{black}\bottomrule
    \end{tabular}}
    \vspace{-2em}
    \label{tab:ab_rldmd}
\end{subtable}
\hfill 
\begin{subtable}{0.44\linewidth}
    \centering  
    \caption{Ablation on Value of $\lambda_{\text{rl}}$.}
    \vspace{-0.5em}
    \resizebox{\linewidth}{!}{
    \begin{tabular}{l c c c c c }
    \toprule
    \multirow{2}{*}{Value} & CLIP & Aesthetic & Pick & HP & DPG\\
      &Score & Score & Score & Score & overall\\
    \arrayrulecolor{black!40}\midrule
    0.1 & 34.7422 & 5.8508 & 22.3064 & 31.6565 & 85.24 \\
    \rowcolor[RGB]{248, 242, 253} 
    0.5 & 35.0142  & 5.8876  &  22.4892 & 32.1145 & 85.30\\
    1.0 & 35.0087 & 5.84430 & 22.4709 & 32.8982 & 85.04 \\
    2.0 & 35.0322 & 5.7806 & 22.1098 & 32.7683 & 82.48 \\
    \arrayrulecolor{black}\bottomrule
    \end{tabular}}
    \vspace{-2em}
    \label{tab:ab_coefficient}
\end{subtable}

\end{table}

\subsection{Ablation Study}
In this section, we conduct a systematic investigation into the key components of \sname . Unless otherwise specified, all experiments use the SD3-M base model with a 4-step sampling configuration.

\noindent\textbf{Versatility across reward models.} A key advantage of ourusing a DINOv2-based reward model to maximize classification accuracy\sname is the ability to steer the distillation process toward diverse human preferences. As shown in Table~\ref{tab:ab_reward_model}, we evaluate the framework using different reward models (RMs) targeting specific attributes. The results provide a clear insight: optimizing with a specific RM leads to a significant performance gain in its corresponding metric, confirming that \sname effectively injects the desired preference signal into the few-step manifold. Furthermore, by integrating multiple rewards (CLIP + HPS), the model achieves a more balanced and superior generative performance, similar to the findings in DanceGRPO~\cite{dancegrpo}. This flexibility allows users to customize the distillation "flavor" according to specific downstream tasks.

\noindent\textbf{Compatibility with RL paradigms.}
To verify the algorithmic universality of our unified framework, we compare three distinct RL strategies in the Stage 2 joint optimization phase: ReFL~\cite{refl}, DPO~\cite{diffusiondpo,pso}, and GRPO~\cite{dancegrpo,flowgrpo}. As shown in Table~\ref{tab:ab_rldmd}, all three algorithms yield substantial improvements over the "Stage 1 init" baseline across all evaluation dimensions. 
Furthermore, an interesting observation is that using the ReFL in \sname achieves overall better performance, and our analysis is that: traditional ReFL for multi-step models ignores earlier sampling steps and only trains the last few steps before the output image~\cite{refl} because the large computational graph generated by  the multi-step denoising process would result in a huge memory consumption. Subsequent solutions~\cite{srpo,deepreward} mainly focus on saving forward noise trajectory and sampling a subset of steps to optimize, although this can allow the gradient to return to the early denoising step, it is also prone to error accumulation. On the contrary, in our few-step model, the model is trained to predict a clean image at each step's state~\cite{dmd2}, which enables the gradient of ReFL to directly return to the initial state, thus obtaining an effective optimization.

\noindent\textbf{The balancing Role of DMD Regularization.}
The synergy between distillation and RL is governed by the balancing coefficient $\lambda_{\text{rl}}$ in our joint objective. Table~\ref{tab:ab_coefficient} illustrates the trade-off between reward maximization and distributional stability. When 
$\lambda_{\text{rl}}$ is set too low; the model remains constrained by the teacher's manifold, limiting the potential for preference alignment. Conversely, as it increases beyond an optimal threshold, we observe a noticeable decline in a general capacity (reflected by the DPG overall score), despite high reward scores. This trend provides a crucial insight: the DMD loss serves as a vital structural regularizer. It prevents the model from "reward hacking"(We also showed in Figure~\ref{fig:verification1}. The optimal balance ensures that the model pushes the boundaries of quality while staying anchored to the teacher's robust generative priors.

\begin{table}[t]
\centering
\vspace{0.5em}
\caption{\textbf{1-step SiT DMDR Results on ImageNet $256\times256$}. We report the performance with different CFG scales and RL optimization. Metrics include FID~\cite{fid}, Recall~\cite{pre-rec}, Inception Score (IS)~\cite{is}, and Precision~\cite{pre-rec}.}
\vspace{-1em}
\label{tab:imagenet_res}
\begin{tabular}{@{}lccccc@{}}
\toprule
CFG &  RL & FID $\downarrow$ & Recall $\uparrow$ & Inception Score $\uparrow$ & Precision $\uparrow$ \\ \midrule
No  & No      & 2.13             & 0.64              & 232.15                     & 0.77                 \\
1.5 & No      & 5.20             & 0.48              & 387.24                     & 0.88                 \\
4   & No      & 15.70            & 0.11              & 453.65                     & 0.86                 \\ \midrule
No  & Yes     & 6.95             & 0.40              & 416.01                     & 0.90                 \\
1.5 & Yes     & 13.43            & 0.24              & 494.43                     & 0.93                 \\ \bottomrule
\end{tabular}
\end{table} 
\section{Limitation and Discussion}

\noindent\textbf{Quality vs. Diversity.} 
To further analyze the behavior of \sname, we conduct class-conditional experiments on ImageNet 256$\times$256
using the SiT~\cite{sit,repa,sra} backbone for 1-step generation and text-to-image settings used in the former paper. As shown in Table~\ref{tab:imagenet_res}, a fundamental trade-off between generative quality and diversity emerges. Our results reveal that both Classifier-Free Guidance (CFG) (this also observed in  Decoupled-DMD~\cite{decoupled-dmd}) and Reinforcement Learning (here we use a DINOv2-based~\cite{dinov2} reward model to maximize classification accuracy) act as "density sharpeners" during distillation.  

We believe this reduction in diversity is not a limitation unique to our approach, but a characteristic of score-based distillation and preference alignment, which steer the model toward high-reward, high-confidence regions. We also provide additional experiments with LPIPS diversity~\cite{lpips} on SD3-medium in Table~\ref{tab:diversity_evaluation}, which shows similar degradation as on ImageNet and for other methods, while DMDR provides a favorable quality/alignment trade-off. Although DMD regularization mitigates reward hacking compared with direct/sequential RL (shown in Figure~\ref{fig:verification1}), it cannot completely remove the inherent diversity cost of reward optimization. This trade-off is tunable: a smaller $\lambda_{\rm rl}$ improves diversity with reduced preference gains as shown in Table~\ref{tab:trade-off}, allowing users to balance the trade-off. More explorations on adaptive reward weighting and other diversity-preserving RL objectives could be  future work.

\begin{table}[t]
\centering
\caption{Diversity evaluation on ShareGPT-4o-Image~\cite{sharegpt4o-image}.}
\vspace{-0.5em}
\begin{tabular}{l|cccccc}
\toprule
\textbf{Metric} & Base-Model & DMD & DMD2 & Base + ReFL & DMD + PSO & DMDR \\
\midrule
HPS$\uparrow$ & 28.4021 & 28.0248 & 27.3675 & 32.0745 & 30.1180 & 32.1145 \\
LPIPS$\uparrow$ & 0.6840 & 0.5664 & 0.5832 & 0.5602 & 0.5540 & 0.5548 \\
\bottomrule
\end{tabular}
\label{tab:diversity_evaluation}
\end{table}

\begin{table}[t]
\centering
\caption{Impact of the trade-off $\lambda_{\rm rl}$.}
\vspace{-0.5em}
\label{tab:trade-off}
\resizebox{0.65\textwidth}{!}{  
\begin{tabular}{l|ccccc}
\toprule
\textbf{Metric} & Base-Model & DMD & DMDR $\lambda_{\rm rl}0.1$ & DMDR $\lambda_{\rm rl}0.5$ & DMDR $\lambda_{\rm rl}1$ \\
\midrule
HPS$\uparrow$ & 28.4021 & 28.0248  & 30.4024 & 32.1145 & 32.4786 \\
LPIPS$\uparrow$ & 0.6840 & 0.5664  & 0.5602 & 0.5548 & 0.5266 \\
\bottomrule
\end{tabular}
}
\end{table}

\section{Related Work}

\noindent\textbf{Distribution matching distillation.}
Distribution Matching Distillation~\cite{dmd} (DMD) is the first work to successfully apply the principle score-based distillation~\cite{score-base-diff,prolificdreamer,dreamfusion} to large-scale text-to-image models. Intuitively, it strives to ensure that any sample realized by the student at a given noise level occurs with exactly the same probability as it would under the teacher’s distribution, thereby preserving the multi-step generative priors in the few-step model. From then on, a lot of follow-up work emerged~\cite{dmd2,dmdx,tdm,sd3.5flash,f-distill,flash-dmd,twinflow}, for example, DMD2~\cite{dmd2} employs a discriminator to align the student model distribution with a specific target distribution like GAN~\cite{gan} does; TDM~\cite{tdm} incorporates DMD loss in the sampling process of the student model for better alignment; $f$--distill~\cite{f-distill} replaces the original reverse Kullback–Leibler (KL) divergence to the proposed $f$-divergence for covering different divergences with different properties. Our work also starts with DMD, but we don't focus on how to better ``imitate" the teacher. Instead, we aim to incorporate reinforcement learning in the distillation process to enable a more controllable and preference-aligned distillation.

\noindent\textbf{Reinforce learning for diffusion models.}
Reinforce learning helps to align diffusion models to human preferences by training a reward model and using it to guide generation~\cite{ddpo,flowgrpo,refl,srpo,diffusiondpo,dancegrpo,diffusionnft}. There are many algorithms to conduct RL, for example, DDPO~\cite{ddpo} adapts PPO via image log-likelihoods; ReFL~\cite{refl} bypasses likelihoods by optimizing outputs with frozen-reward gradients; Diffusion-DPO~\cite{diffusiondpo} adapts DPO to diffusion for paired human preference
data; and recent GRPO extensions~\cite{flowgrpo,dancegrpo} use GRPO to diffusion models by coupling the training
loss with SDE samplers. Although significant progress has been made in RL for diffusion models, most of the work has focused on multi-step models. Here, we find that these algorithms are also adapted to the few-step model when carried out in conjunction with distribution matching distillation.

\noindent\textbf{Trajectory-based distillation.} Trajectory-based distillation~\cite{lcm,hypersd,rcm,piflow}, which typically aims to simulate teacher ODE trajectories on the instance level. Early work \cite{kdgen} regresses the teacher's ODE integral in one step, producing blurry $\ell_2$--$x_0$ estimates, and thus suffers from degraded quality. Progressive distillation \cite{hypersd,shortcut-model} mitigates this by a multi-stage pipeline that enlarges the student’s step size and halves its NFE each stage by distilling the prior stage’s trajectory into fewer steps, However, this not only makes the training a multi-stage process (e.g., NFE from 50 to 25 then to 10, etc.), but also leads to cumulative errors. Meanwhile, consistency models~\cite{lcm,scm} aim to learn a consistency function that maps the point at an arbitrary time t on the teacher's PF-ODE trajectory to the initial point. However, the student model must be constructed implicitly using either inaccurate finite differences or expensive Jacobian–vector products (JVPs) and the quality is still limited due to the accumulation of errors into the integrated state, which limits the application in large-scale scenarios~\cite{rcm,piflow}.

\noindent\textbf{Adversarial distillation.} Adversarial distillation~\cite{add,apt,ladd,apt2} can be regarded as another form of distribution matching distillation. The difference is that adversarial distillation aim to estimate the distribution of both the student model and the real data or the teacher model through a discriminator model~\cite{gan} to match the distribution. Representative works like Adversarial Diffusion Distillation (ADD)~\cite{add} force the few-step student diffusion model to fool a discriminator which is trained to distinguish the generated samples from real images. Follow-up works~\cite{apt,apt2} applied this idea to other fields. However, such a training method can cause many instabilities, similar to traditional GANs~\cite{gan,improvedgan}, hence requires a lot of tricks to stabilize training~\cite{animatediff-l,r1reg,train-gan}.

\noindent\textbf{Other distillation works.} Other distillation works mainly focus on combining the distillation at the distribution level and the trajectory level~\cite{sana-sprint,rcm,tdm,pose}. For example, SANA-Sprint~\cite{sana-sprint} combines the ideas of LADD~\cite{ladd} and sCM~\cite{scm} and achieves fast convergence and high fidelity generation while retaining the alignment advantages of sCMs. TDM~\cite{tdm} and rCM~\cite{rcm} combine DMD~\cite{dmd} and LCM~\cite{lcm} together to remedy the quality issues of LCM.

\noindent\textbf{Reward models for diffusion post-train.} Unlike Large Language Models (LLMs), where reinforcement learning (RL) can often leverage verifiable, rule-based rewards~\cite{deepseekmath,deepseekr1,glm} (e.g., code execution success or mathematical correctness), diffusion-based image generation is inherently subjective. The evaluation of generated images relies heavily on multifaceted criteria such as aesthetic quality, instruction following, and so on, which are difficult to quantify through rigid, rule-based functions. Consequently, the development of robust reward models (RMs) has become a critical prerequisite for effective RL-based alignment in diffusion models~\cite{refl,rewarddance,clipscore,hps,visionreward,q-align}.
Early methods, such as CLIP-score~\cite{clipscore}, primarily focus on measuring the semantic alignment between the text prompt and the generated image. To address the need for visual fidelity and artistic appeal, metrics like Aesthetic Score~\cite{laion5b} and HPS (Human Preference Score)~\cite{hps,hpsv2,hpsv3} have been introduced to capture the nuance of human aesthetic judgment. More recently, the reward models become unified~\cite{unified-reward,visionreward,unified-p-reward} and aggregate multi-dimensional preferences—ranging from low-level texture details to high-level semantic instruction following—into a unified scoring function. While our primary contribution lies in proposing a novel algorithm that integrates few-step distillation with reinforcement learning, our empirical results highlight the pivotal role of the reward model in our pipeline.

\section{Conclusion}
We introduces \sname, a unified framework that transforms Distribution Matching Distillation (DMD) and Reinforcement Learning (RL) from independent, sequential stages into a synergistic training pipeline. We demonstrate that joint optimization is mutually beneficial: RL steers distillation toward high-reward regions to break the "teacher ceiling," while DMD regularizes RL to effectively mitigate reward hacking. By employing a two-stage strategy—incorporating RT-DM and dynamic training followed by joint optimization—\sname achieves state-of-the-art few-step performance across diverse architectures. Ultimately, our approach enables few-step models to not only match but consistently surpass the visual quality and prompt adherence of their original multi-step teacher models.

\clearpage
{
    \small
    \bibliographystyle{ieeenat_fullname}
    \bibliography{main}

@String(TOG= {ACM Trans. Graph.})

@String(ICLR = {Int. Conf. Learn. Represent.})

@String(AAAI = {AAAI})

@String(TOG   = {ACM TOG})

@String(ICLR  = {ICLR})

@inproceedings{dmd,
  title={One-step diffusion with distribution matching distillation},
  author={Yin, Tianwei and Gharbi, Micha{\"e}l and Zhang, Richard and Shechtman, Eli and Durand, Fredo and Freeman, William T and Park, Taesung},
  booktitle={Proceedings of the IEEE/CVF conference on computer vision and pattern recognition},
  pages={6613--6623},
  year={2024}
}

@article{dmd2,
  title={Improved distribution matching distillation for fast image synthesis},
  author={Yin, Tianwei and Gharbi, Micha{\"e}l and Park, Taesung and Zhang, Richard and Shechtman, Eli and Durand, Fredo and Freeman, Bill},
  journal={Advances in neural information processing systems},
  volume={37},
  pages={47455--47487},
  year={2024}
}

@inproceedings{ladd,
  title={Fast high-resolution image synthesis with latent adversarial diffusion distillation},
  author={Sauer, Axel and Boesel, Frederic and Dockhorn, Tim and Blattmann, Andreas and Esser, Patrick and Rombach, Robin},
  booktitle={SIGGRAPH Asia 2024 Conference Papers},
  pages={1--11},
  year={2024}
}

@inproceedings{dit,
  title={Scalable diffusion models with transformers},
  author={Peebles, William and Xie, Saining},
  booktitle={Proceedings of the IEEE/CVF international conference on computer vision},
  pages={4195--4205},
  year={2023}
}

@misc{flux,
    author={Black Forest Labs},
    title={FLUX},
    year={2024},
    howpublished={\url{https://github.com/black-forest-labs/flux}},
}

@inproceedings{sd3,
  title={Scaling rectified flow transformers for high-resolution image synthesis},
  author={Esser, Patrick and Kulal, Sumith and Blattmann, Andreas and Entezari, Rahim and M{\"u}ller, Jonas and Saini, Harry and Levi, Yam and Lorenz, Dominik and Sauer, Axel and Boesel, Frederic and others},
  booktitle={Forty-first international conference on machine learning},
  year={2024}
}

@article{hypersd,
  title={Hyper-sd: Trajectory segmented consistency model for efficient image synthesis},
  author={Ren, Yuxi and Xia, Xin and Lu, Yanzuo and Zhang, Jiacheng and Wu, Jie and Xie, Pan and Wang, Xing and Xiao, Xuefeng},
  journal={arXiv preprint arXiv:2404.13686},
  year={2024}
}

@article{gan,
  title={Generative adversarial nets},
  author={Goodfellow, Ian and Pouget-Abadie, Jean and Mirza, Mehdi and Xu, Bing and Warde-Farley, David and Ozair, Sherjil and Courville, Aaron and Bengio, Yoshua},
  journal={Advances in neural information processing systems},
  volume={27},
  year={2014}
}

@article{seedream4,
  title={Seedream 4.0: Toward next-generation multimodal image generation},
  author={Seedream, Team and Chen, Yunpeng and Gao, Yu and Gong, Lixue and Guo, Meng and Guo, Qiushan and Guo, Zhiyao and Hou, Xiaoxia and Huang, Weilin and Huang, Yixuan and others},
  journal={arXiv preprint arXiv:2509.20427},
  year={2025}
}

@article{tdm,
  title={Learning Few-Step Diffusion Models by Trajectory Distribution Matching},
  author={Luo, Yihong and Hu, Tianyang and Sun, Jiacheng and Cai, Yujun and Tang, Jing},
  journal={arXiv preprint arXiv:2503.06674},
  year={2025}
}

@article{dmdx,
  title={Adversarial distribution matching for diffusion distillation towards efficient image and video synthesis},
  author={Lu, Yanzuo and Ren, Yuxi and Xia, Xin and Lin, Shanchuan and Wang, Xing and Xiao, Xuefeng and Ma, Andy J and Xie, Xiaohua and Lai, Jian-Huang},
  journal={arXiv preprint arXiv:2507.18569},
  year={2025}
}

@article{diffinstruct,
  title={Diff-instruct: A universal approach for transferring knowledge from pre-trained diffusion models},
  author={Luo, Weijian and Hu, Tianyang and Zhang, Shifeng and Sun, Jiacheng and Li, Zhenguo and Zhang, Zhihua},
  journal={Advances in Neural Information Processing Systems},
  volume={36},
  pages={76525--76546},
  year={2023}
}

@article{diffinstruct++,
  title={Diff-instruct++: Training one-step text-to-image generator model to align with human preferences},
  author={Luo, Weijian},
  journal={arXiv preprint arXiv:2410.18881},
  year={2024}}

@article{lcm,
  title={Latent consistency models: Synthesizing high-resolution images with few-step inference},
  author={Luo, Simian and Tan, Yiqin and Huang, Longbo and Li, Jian and Zhao, Hang},
  journal={arXiv preprint arXiv:2310.04378},
  year={2023}}

@inproceedings{flashdiff,
  title={Flash diffusion: Accelerating any conditional diffusion model for few steps image generation},
  author={Chadebec, Clement and Tasar, Onur and Benaroche, Eyal and Aubin, Benjamin},
  booktitle={Proceedings of the AAAI Conference on Artificial Intelligence},
  volume={39},
  number={15},
  pages={15686--15695},
  year={2025}
}

@article{magicdistillation,
  title={MagicDistillation: Weak-to-Strong Video Distillation for Large-Scale Few-Step Synthesis},
  author={Shao, Shitong and Yi, Hongwei and Guo, Hanzhong and Ye, Tian and Zhou, Daquan and Lingelbach, Michael and Xu, Zhiqiang and Xie, Zeke},
  journal={arXiv preprint arXiv:2503.13319},
  year={2025}
}

@article{sd3.5flash,
  title={SD3. 5-Flash: Distribution-Guided Distillation of Generative Flows},
  author={Bandyopadhyay, Hmrishav and Entezari, Rahim and Scott, Jim and Adithyan, Reshinth and Song, Yi-Zhe and Jampani, Varun},
  journal={arXiv preprint arXiv:2509.21318},
  year={2025}
}

@article{refl,
  title={Imagereward: Learning and evaluating human preferences for text-to-image generation},
  author={Xu, Jiazheng and Liu, Xiao and Wu, Yuchen and Tong, Yuxuan and Li, Qinkai and Ding, Ming and Tang, Jie and Dong, Yuxiao},
  journal={Advances in Neural Information Processing Systems},
  volume={36},
  pages={15903--15935},
  year={2023}
}

@article{srpo,
  title={Directly aligning the full diffusion trajectory with fine-grained human preference},
  author={Shen, Xiangwei and Li, Zhimin and Yang, Zhantao and Zhang, Shiyi and Zhang, Yingfang and Li, Donghao and Wang, Chunyu and Lu, Qinglin and Tang, Yansong},
  journal={arXiv preprint arXiv:2509.06942},
  year={2025}
}

@article{flowgrpo,
  title={Flow-grpo: Training flow matching models via online rl},
  author={Liu, Jie and Liu, Gongye and Liang, Jiajun and Li, Yangguang and Liu, Jiaheng and Wang, Xintao and Wan, Pengfei and Zhang, Di and Ouyang, Wanli},
  journal={arXiv preprint arXiv:2505.05470},
  year={2025}
}

@article{dancegrpo,
  title={DanceGRPO: Unleashing GRPO on Visual Generation},
  author={Xue, Zeyue and Wu, Jie and Gao, Yu and Kong, Fangyuan and Zhu, Lingting and Chen, Mengzhao and Liu, Zhiheng and Liu, Wei and Guo, Qiushan and Huang, Weilin and others},
  journal={arXiv preprint arXiv:2505.07818},
  year={2025}
}

@article{rcm,
  title={Large Scale Diffusion Distillation via Score-Regularized Continuous-Time Consistency},
  author={Zheng, Kaiwen and Wang, Yuji and Ma, Qianli and Chen, Huayu and Zhang, Jintao and Balaji, Yogesh and Chen, Jianfei and Liu, Ming-Yu and Zhu, Jun and Zhang, Qinsheng},
  journal={arXiv preprint arXiv:2510.08431},
  year={2025}
}

@article{improvedgan,
  title={Improved techniques for training gans},
  author={Salimans, Tim and Goodfellow, Ian and Zaremba, Wojciech and Cheung, Vicki and Radford, Alec and Chen, Xi},
  journal={Advances in neural information processing systems},
  volume={29},
  year={2016}
}

@article{sdxl,
  title={Sdxl: Improving latent diffusion models for high-resolution image synthesis},
  author={Podell, Dustin and English, Zion and Lacey, Kyle and Blattmann, Andreas and Dockhorn, Tim and M{\"u}ller, Jonas and Penna, Joe and Rombach, Robin},
  journal={arXiv preprint arXiv:2307.01952},
  year={2023}
}

@article{f-distill,
  title={One-step Diffusion Models with $ f $-Divergence Distribution Matching},
  author={Xu, Yilun and Nie, Weili and Vahdat, Arash},
  journal={arXiv preprint arXiv:2502.15681},
  year={2025}
}

@article{dreamfusion,
  title={Dreamfusion: Text-to-3d using 2d diffusion},
  author={Poole, Ben and Jain, Ajay and Barron, Jonathan T and Mildenhall, Ben},
  journal={arXiv preprint arXiv:2209.14988},
  year={2022}
}

@article{prolificdreamer,
  title={Prolificdreamer: High-fidelity and diverse text-to-3d generation with variational score distillation},
  author={Wang, Zhengyi and Lu, Cheng and Wang, Yikai and Bao, Fan and Li, Chongxuan and Su, Hang and Zhu, Jun},
  journal={Advances in neural information processing systems},
  volume={36},
  pages={8406--8441},
  year={2023}
}

@article{ddpo,
  title={Training diffusion models with reinforcement learning},
  author={Black, Kevin and Janner, Michael and Du, Yilun and Kostrikov, Ilya and Levine, Sergey},
  journal={arXiv preprint arXiv:2305.13301},
  year={2023}
}

@article{score-base-diff,
  title={Score-based generative modeling through stochastic differential equations},
  author={Song, Yang and Sohl-Dickstein, Jascha and Kingma, Diederik P and Kumar, Abhishek and Ermon, Stefano and Poole, Ben},
  journal={arXiv preprint arXiv:2011.13456},
  year={2020}
}

@article{dpg,
  title={Ella: Equip diffusion models with llm for enhanced semantic alignment},
  author={Hu, Xiwei and Wang, Rui and Fang, Yixiao and Fu, Bin and Cheng, Pei and Yu, Gang},
  journal={arXiv preprint arXiv:2403.05135},
  year={2024}
}

@article{geneval,
  title={Geneval: An object-focused framework for evaluating text-to-image alignment},
  author={Ghosh, Dhruba and Hajishirzi, Hannaneh and Schmidt, Ludwig},
  journal={Advances in Neural Information Processing Systems},
  volume={36},
  pages={52132--52152},
  year={2023}
}

@article{sharegpt4o-image,
  title={ShareGPT-4o-Image: Aligning Multimodal Models with GPT-4o-Level Image Generation},
  author={Chen, Junying and Cai, Zhenyang and Chen, Pengcheng and Chen, Shunian and Ji, Ke and Wang, Xidong and Yang, Yunjin and Wang, Benyou},
  journal={arXiv preprint arXiv:2506.18095},
  year={2025}
}

@article{clipscore,
  title={Clipscore: A reference-free evaluation metric for image captioning},
  author={Hessel, Jack and Holtzman, Ari and Forbes, Maxwell and Bras, Ronan Le and Choi, Yejin},
  journal={arXiv preprint arXiv:2104.08718},
  year={2021}
}

@article{pickscore,
  title={Pick-a-pic: An open dataset of user preferences for text-to-image generation},
  author={Kirstain, Yuval and Polyak, Adam and Singer, Uriel and Matiana, Shahbuland and Penna, Joe and Levy, Omer},
  journal={Advances in neural information processing systems},
  volume={36},
  pages={36652--36663},
  year={2023}
}

@article{laion5b,
  title={Laion-5b: An open large-scale dataset for training next generation image-text models},
  author={Schuhmann, Christoph and Beaumont, Romain and Vencu, Richard and Gordon, Cade and Wightman, Ross and Cherti, Mehdi and Coombes, Theo and Katta, Aarush and Mullis, Clayton and Wortsman, Mitchell and others},
  journal={Advances in neural information processing systems},
  volume={35},
  pages={25278--25294},
  year={2022}
}

@article{hpsv2,
  title={Human preference score v2: A solid benchmark for evaluating human preferences of text-to-image synthesis},
  author={Wu, Xiaoshi and Hao, Yiming and Sun, Keqiang and Chen, Yixiong and Zhu, Feng and Zhao, Rui and Li, Hongsheng},
  journal={arXiv preprint arXiv:2306.09341},
  year={2023}
}

@misc{sd35,
    author={Stability AI.},
    title={Sd3.5},
    year={2024},
    howpublished={\url{https://github.com/Stability-AI/sd3.5}},
}

@article{dfnclip,
  title={Data Filtering Networks},
  author={Fang, Alex and Jose, Albin Madappally and Jain, Amit and Schmidt, Ludwig and Toshev, Alexander and Shankar, Vaishaal},
  journal={arXiv preprint arXiv:2309.17425},
  year={2023}
}

@article{sdxl-lightning,
  title={Sdxl-lightning: Progressive adversarial diffusion distillation},
  author={Lin, Shanchuan and Wang, Anran and Yang, Xiao},
  journal={arXiv preprint arXiv:2402.13929},
  year={2024}
}

@article{lora,
  title={Lora: Low-rank adaptation of large language models.},
  author={Hu, Edward J and Shen, Yelong and Wallis, Phillip and Allen-Zhu, Zeyuan and Li, Yuanzhi and Wang, Shean and Wang, Lu and Chen, Weizhu and others},
  journal={ICLR},
  volume={1},
  number={2},
  pages={3},
  year={2022}
}

@article{scm,
  title={Simplifying, stabilizing and scaling continuous-time consistency models},
  author={Lu, Cheng and Song, Yang},
  journal={arXiv preprint arXiv:2410.11081},
  year={2024}
}

@article{sana-sprint,
  title={Sana-sprint: One-step diffusion with continuous-time consistency distillation},
  author={Chen, Junsong and Xue, Shuchen and Zhao, Yuyang and Yu, Jincheng and Paul, Sayak and Chen, Junyu and Cai, Han and Han, Song and Xie, Enze},
  journal={arXiv preprint arXiv:2503.09641},
  year={2025}
}

@article{piflow,
  title={pi-Flow: Policy-Based Few-Step Generation via Imitation Distillation},
  author={Chen, Hansheng and Zhang, Kai and Tan, Hao and Guibas, Leonidas and Wetzstein, Gordon and Bi, Sai},
  journal={arXiv preprint arXiv:2510.14974},
  year={2025}
}

@article{kdgen,
  title={Knowledge distillation in iterative generative models for improved sampling speed},
  author={Luhman, Eric and Luhman, Troy},
  journal={arXiv preprint arXiv:2101.02388},
  year={2021}
}

@article{shortcut-model,
  title={One step diffusion via shortcut models},
  author={Frans, Kevin and Hafner, Danijar and Levine, Sergey and Abbeel, Pieter},
  journal={arXiv preprint arXiv:2410.12557},
  year={2024}
}

@article{apt2,
  title={Autoregressive Adversarial Post-Training for Real-Time Interactive Video Generation},
  author={Lin, Shanchuan and Yang, Ceyuan and He, Hao and Jiang, Jianwen and Ren, Yuxi and Xia, Xin and Zhao, Yang and Xiao, Xuefeng and Jiang, Lu},
  journal={arXiv preprint arXiv:2506.09350},
  year={2025}
}

@article{apt,
  title={Diffusion adversarial post-training for one-step video generation},
  author={Lin, Shanchuan and Xia, Xin and Ren, Yuxi and Yang, Ceyuan and Xiao, Xuefeng and Jiang, Lu},
  journal={arXiv preprint arXiv:2501.08316},
  year={2025}
}

@inproceedings{add,
  title={Adversarial diffusion distillation},
  author={Sauer, Axel and Lorenz, Dominik and Blattmann, Andreas and Rombach, Robin},
  booktitle={European Conference on Computer Vision},
  pages={87--103},
  year={2024},
  organization={Springer}
}

@article{animatediff-l,
  title={Animatediff-lightning: Cross-model diffusion distillation},
  author={Lin, Shanchuan and Yang, Xiao},
  journal={arXiv preprint arXiv:2403.12706},
  year={2024}
}

@article{r1reg,
  title={Stabilizing training of generative adversarial networks through regularization},
  author={Roth, Kevin and Lucchi, Aurelien and Nowozin, Sebastian and Hofmann, Thomas},
  journal={Advances in neural information processing systems},
  volume={30},
  year={2017}
}

@inproceedings{train-gan,
  title={Which training methods for GANs do actually converge?},
  author={Mescheder, Lars and Geiger, Andreas and Nowozin, Sebastian},
  booktitle={International conference on machine learning},
  pages={3481--3490},
  year={2018},
  organization={PMLR}
}

@article{pose,
  title={Pose: Phased one-step adversarial equilibrium for video diffusion models},
  author={Cheng, Jiaxiang and Ma, Bing and Ren, Xuhua and Jin, Hongyi and Yu, Kai and Zhang, Peng and Li, Wenyue and Zhou, Yuan and Zheng, Tianxiang and Lu, Qinglin},
  journal={arXiv preprint arXiv:2508.21019},
  year={2025}
}

@article{lsast,
  title={Towards highly realistic artistic style transfer via stable diffusion with step-aware and layer-aware prompt},
  author={Zhang, Zhanjie and Zhang, Quanwei and Lin, Huaizhong and Xing, Wei and Mo, Juncheng and Huang, Shuaicheng and Xie, Jinheng and Li, Guangyuan and Luan, Junsheng and Zhao, Lei and others},
  journal={arXiv preprint arXiv:2404.11474},
  year={2024}
}

@article{prospect,
  title={Prospect: Prompt spectrum for attribute-aware personalization of diffusion models},
  author={Zhang, Yuxin and Dong, Weiming and Tang, Fan and Huang, Nisha and Huang, Haibin and Ma, Chongyang and Lee, Tong-Yee and Deussen, Oliver and Xu, Changsheng},
  journal={ACM Transactions on Graphics (TOG)},
  volume={42},
  number={6},
  pages={1--14},
  year={2023},
  publisher={ACM New York, NY, USA}
}

@article{hypernoise,
  title={Noise Hypernetworks: Amortizing Test-Time Compute in Diffusion Models},
  author={Eyring, Luca and Karthik, Shyamgopal and Dosovitskiy, Alexey and Ruiz, Nataniel and Akata, Zeynep},
  journal={arXiv preprint arXiv:2508.09968},
  year={2025}
}

@article{rewarddance,
  title={Rewarddance: Reward scaling in visual generation},
  author={Wu, Jie and Gao, Yu and Ye, Zilyu and Li, Ming and Li, Liang and Guo, Hanzhong and Liu, Jie and Xue, Zeyue and Hou, Xiaoxia and Liu, Wei and others},
  journal={arXiv preprint arXiv:2509.08826},
  year={2025}
}

@inproceedings{deepreward,
  title={Deep reward supervisions for tuning text-to-image diffusion models},
  author={Wu, Xiaoshi and Hao, Yiming and Zhang, Manyuan and Sun, Keqiang and Huang, Zhaoyang and Song, Guanglu and Liu, Yu and Li, Hongsheng},
  booktitle={European Conference on Computer Vision},
  pages={108--124},
  year={2024},
  organization={Springer}
}

@inproceedings{lpips,
  title={The unreasonable effectiveness of deep features as a perceptual metric},
  author={Zhang, Richard and Isola, Phillip and Efros, Alexei A and Shechtman, Eli and Wang, Oliver},
  booktitle={Proceedings of the IEEE conference on computer vision and pattern recognition},
  pages={586--595},
  year={2018}
}

@article{flow-dpo,
  title={Improving video generation with human feedback},
  author={Liu, Jie and Liu, Gongye and Liang, Jiajun and Yuan, Ziyang and Liu, Xiaokun and Zheng, Mingwu and Wu, Xiele and Wang, Qiulin and Qin, Wenyu and Xia, Menghan and others},
  journal={arXiv preprint arXiv:2501.13918},
  year={2025}
}

@article{zimage,
  title={Z-Image: An Efficient Image Generation Foundation Model with Single-Stream Diffusion Transformer},
  author={Z-Image Team},
  journal={arXiv preprint arXiv:2511.22699},
  year={2025}
}

@article{pso,
  title={Tuning Timestep-Distilled Diffusion Model Using Pairwise Sample Optimization},
  author={Miao, Zichen and Yang, Zhengyuan and Lin, Kevin and Wang, Ze and Liu, Zicheng and Wang, Lijuan and Qiu, Qiang},
  journal={arXiv preprint arXiv:2410.03190},
  year={2024}
}

@article{diffusionnft,
  title={DiffusionNFT: Online Diffusion Reinforcement with Forward Process},
  author={Zheng, Kaiwen and Chen, Huayu and Ye, Haotian and Wang, Haoxiang and Zhang, Qinsheng and Jiang, Kai and Su, Hang and Ermon, Stefano and Zhu, Jun and Liu, Ming-Yu},
  journal={arXiv preprint arXiv:2509.16117},
  year={2025}
}

@article{flash-dmd,
  title={Flash-DMD: Towards High-Fidelity Few-Step Image Generation with Efficient Distillation and Joint Reinforcement Learning},
  author={Guanjie Chen and Shirui Huang and Kai Liu and Jian-Xiang Zhu and Xiaoye Qu and Peng Chen and Yu Cheng and Yifu Sun},
  journal={ArXiv},
  year={2025},
  volume={abs/2511.20549}
}

@inproceedings{diffusiondpo,
  title={Diffusion model alignment using direct preference optimization},
  author={Wallace, Bram and Dang, Meihua and Rafailov, Rafael and Zhou, Linqi and Lou, Aaron and Purushwalkam, Senthil and Ermon, Stefano and Xiong, Caiming and Joty, Shafiq and Naik, Nikhil},
  booktitle={Proceedings of the IEEE/CVF Conference on Computer Vision and Pattern Recognition},
  pages={8228--8238},
  year={2024}
}

@article{ddim,
  title={Denoising diffusion implicit models},
  author={Song, Jiaming and Meng, Chenlin and Ermon, Stefano},
  journal={arXiv preprint arXiv:2010.02502},
  year={2020}
}

@article{ddpm,
  title={Denoising diffusion probabilistic models},
  author={Ho, Jonathan and Jain, Ajay and Abbeel, Pieter},
  journal={Advances in neural information processing systems},
  volume={33},
  pages={6840--6851},
  year={2020}
}

@article{flow-matching,
  title={Flow matching for generative modeling},
  author={Lipman, Yaron and Chen, Ricky TQ and Ben-Hamu, Heli and Nickel, Maximilian and Le, Matt},
  journal={arXiv preprint arXiv:2210.02747},
  year={2022}
}

@inproceedings{hpsv3,
  title={Hpsv3: Towards wide-spectrum human preference score},
  author={Ma, Yuhang and Wu, Xiaoshi and Sun, Keqiang and Li, Hongsheng},
  booktitle={Proceedings of the IEEE/CVF International Conference on Computer Vision},
  pages={15086--15095},
  year={2025}
}

@misc{flux-2,
    author={Black Forest Labs},
    title={{FLUX.2: Frontier Visual Intelligence}},
    year={2025},
    howpublished={\url{https://bfl.ai/blog/flux-2}},
}

@article{senseflow,
  title={SenseFlow: Scaling Distribution Matching for Flow-based Text-to-Image Distillation},
  author={Ge, Xingtong and Zhang, Xin and Xu, Tongda and Zhang, Yi and Zhang, Xinjie and Wang, Yan and Zhang, Jun},
  journal={arXiv preprint arXiv:2506.00523},
  year={2025}
}

@article{twinflow,
  title={TwinFlow: Realizing One-step Generation on Large Models with Self-adversarial Flows},
  author={Cheng, Zhenglin and Sun, Peng and Li, Jianguo and Lin, Tao},
  journal={arXiv preprint arXiv:2512.05150},
  year={2025}
}

@article{fid,
  title={Gans trained by a two time-scale update rule converge to a local nash equilibrium},
  author={Heusel, Martin and Ramsauer, Hubert and Unterthiner, Thomas and Nessler, Bernhard and Hochreiter, Sepp},
  journal={Advances in neural information processing systems},
  volume={30},
  year={2017}
}

@article{pre-rec,
  title={Improved precision and recall metric for assessing generative models},
  author={Kynk{\"a}{\"a}nniemi, Tuomas and Karras, Tero and Laine, Samuli and Lehtinen, Jaakko and Aila, Timo},
  journal={Advances in neural information processing systems},
  volume={32},
  year={2019}
}

@article{is,
  title={Improved techniques for training gans},
  author={Salimans, Tim and Goodfellow, Ian and Zaremba, Wojciech and Cheung, Vicki and Radford, Alec and Chen, Xi},
  journal={Advances in neural information processing systems},
  volume={29},
  year={2016}
}

@inproceedings{sit,
  title={Sit: Exploring flow and diffusion-based generative models with scalable interpolant transformers},
  author={Ma, Nanye and Goldstein, Mark and Albergo, Michael S and Boffi, Nicholas M and Vanden-Eijnden, Eric and Xie, Saining},
  booktitle={European Conference on Computer Vision},
  pages={23--40},
  year={2024},
  organization={Springer}
}

@article{dinov2,
  title={Dinov2: Learning robust visual features without supervision},
  author={Maxime Oquab and Timoth{\'e}e Darcet and Th{\'e}o Moutakanni and Huy V. Vo and Marc Szafraniec and Vasil Khalidov and Pierre Fernandez and Daniel Haziza and Francisco Massa and Alaaeldin El-Nouby and Mahmoud Assran and Nicolas Ballas and Wojciech Galuba and Russ Howes and Po-Yao (Bernie) Huang and Shang-Wen Li and Ishan Misra and Michael G. Rabbat and Vasu Sharma and Gabriel Synnaeve and Hu Xu and Herv{\'e} J{\'e}gou and Julien Mairal and Patrick Labatut and Armand Joulin and Piotr Bojanowski},
  journal={arXiv preprint arXiv:2304.07193},
  year={2023}
}

@inproceedings{repa,
  title={Representation Alignment for Generation: Training Diffusion Transformers Is Easier Than You Think},
  author={Sihyun Yu and Sangkyung Kwak and Huiwon Jang and Jongheon Jeong and Jonathan Huang and Jinwoo Shin and Saining Xie},
  year={2025},
  booktitle={International Conference on Learning Representations},
}

@article{sra,
  title={No other representation component is needed: Diffusion transformers can provide representation guidance by themselves},
  author={Jiang, Dengyang and Wang, Mengmeng and Li, Liuzhuozheng and Zhang, Lei and Wang, Haoyu and Wei, Wei and Dai, Guang and Zhang, Yanning and Wang, Jingdong},
  journal={arXiv preprint arXiv:2505.02831},
  year={2025}
}

@article{decoupled-dmd,
  title={Decoupled DMD: CFG Augmentation as the Spear, Distribution Matching as the Shield},
  author={Liu, Dongyang and Gao, Peng and Liu, David and Du, Ruoyi and Li, Zhen and Wu, Qilong and Jin, Xin and Cao, Sihan and Zhang, Shifeng and Li, Hongsheng and others},
  journal={arXiv preprint arXiv:2511.22677},
  year={2025}
}

@article{rlhf,
  title={Fine-tuning language models from human preferences},
  author={Ziegler, Daniel M and Stiennon, Nisan and Wu, Jeffrey and Brown, Tom B and Radford, Alec and Amodei, Dario and Christiano, Paul and Irving, Geoffrey},
  journal={arXiv preprint arXiv:1909.08593},
  year={2019}
}

@article{ppo,
  title={Proximal policy optimization algorithms},
  author={Schulman, John and Wolski, Filip and Dhariwal, Prafulla and Radford, Alec and Klimov, Oleg},
  journal={arXiv preprint arXiv:1707.06347},
  year={2017}
}

@article{deepseekmath,
  title={Deepseekmath: Pushing the limits of mathematical reasoning in open language models},
  author={Shao, Zhihong and Wang, Peiyi and Zhu, Qihao and Xu, Runxin and Song, Junxiao and Bi, Xiao and Zhang, Haowei and Zhang, Mingchuan and Li, YK and Wu, Yang and others},
  journal={arXiv preprint arXiv:2402.03300},
  year={2024}
}

@article{deepseekr1,
  title={Deepseek-r1: Incentivizing reasoning capability in llms via reinforcement learning},
  author={Guo, Daya and Yang, Dejian and Zhang, Haowei and Song, Junxiao and Wang, Peiyi and Zhu, Qihao and Xu, Runxin and Zhang, Ruoyu and Ma, Shirong and Bi, Xiao and others},
  journal={arXiv preprint arXiv:2501.12948},
  year={2025}
}

@article{glm,
  title={GLM-5: from Vibe Coding to Agentic Engineering},
  author={Zeng, Aohan and Lv, Xin and Hou, Zhenyu and Du, Zhengxiao and Zheng, Qinkai and Chen, Bin and Yin, Da and Ge, Chendi and Xie, Chengxing and Wang, Cunxiang and others},
  journal={arXiv preprint arXiv:2602.15763},
  year={2026}
}

@article{visionreward,
  title={Visionreward: Fine-grained multi-dimensional human preference learning for image and video generation},
  author={Xu, Jiazheng and Huang, Yu and Cheng, Jiale and Yang, Yuanming and Xu, Jiajun and Wang, Yuan and Duan, Wenbo and Yang, Shen and Jin, Qunlin and Li, Shurun and others},
  journal={arXiv preprint arXiv:2412.21059},
  year={2024}
}

@article{q-align,
  title={Q-align: Teaching lmms for visual scoring via discrete text-defined levels},
  author={Wu, Haoning and Zhang, Zicheng and Zhang, Weixia and Chen, Chaofeng and Liao, Liang and Li, Chunyi and Gao, Yixuan and Wang, Annan and Zhang, Erli and Sun, Wenxiu and others},
  journal={arXiv preprint arXiv:2312.17090},
  year={2023}
}

@article{unified-reward,
  title={Unified reward model for multimodal understanding and generation},
  author={Wang, Yibin and Zang, Yuhang and Li, Hao and Jin, Cheng and Wang, Jiaqi},
  journal={arXiv preprint arXiv:2503.05236},
  year={2025}
}

@inproceedings{hps,
  title={Human preference score: Better aligning text-to-image models with human preference},
  author={Wu, Xiaoshi and Sun, Keqiang and Zhu, Feng and Zhao, Rui and Li, Hongsheng},
  booktitle={Proceedings of the IEEE/CVF International Conference on Computer Vision},
  pages={2096--2105},
  year={2023}
}

@article{unified-p-reward,
  title={Unified Personalized Reward Model for Vision Generation},
  author={Wang, Yibin and Zang, Yuhang and Han, Feng and Bu, Jiazi and Zhou, Yujie and Jin, Cheng and Wang, Jiaqi},
  journal={arXiv preprint arXiv:2602.02380},
  year={2026}}
}

\end{document}